\documentclass[11pt]{article}

\usepackage[final]{acl}

\usepackage{times}
\usepackage{latexsym}

\usepackage[T1]{fontenc}

\usepackage[utf8]{inputenc}

\usepackage{microtype}

\usepackage{inconsolata}

\usepackage{graphicx}
\usepackage{booktabs}
\usepackage{multirow}
\usepackage{float}
\usepackage{tikz}
\usetikzlibrary{arrows.meta, positioning, shapes.geometric}
\usepackage{pgfplots}
\pgfplotsset{compat=1.17}
\usepackage{xcolor}
\usepackage{colortbl}   
\usepackage{amsmath}

\title{Lost in Translation, Found in Other Benchmarks: \\ Cross-Task Transfer for Natural Language Inference}

\author{
Ron Keinan \and
Reut Tsarfaty \\
Bar-Ilan University \\
\texttt{\{ronke21, reut.tsarfaty\}@gmail.com}
}

\newcommand{\ignore}[1]{}

\begin{document}
\maketitle
\begin{abstract}
Natural language inference (NLI) is widely used to evaluate language understanding, yet evaluation in many mid- and low-resource languages still relies on benchmarks translated from English. Although translation enables cross-lingual comparison, it may fail to reflect the native linguistic and cultural nuances.
We propose an alternative: construct NLI benchmarks by repurposing existing annotated tasks in the target language, rather than translating English NLI examples. Our pipeline synthesizes premise–hypothesis pairs from native task annotations, generates rule-based contradictions filtered by an LLM, and selects neutral pairs using semantic similarity.
Applying this approach to five existing Hebrew datasets, we introduce \textsc{HEntailment}, a native Hebrew NLI benchmark comprising 161K training and development pairs and 12.6K test pairs. 
We use it to examine transfer between native and translated NLI data in both directions. Fine-tuning on \textsc{HEntailment} improves performance on translated HebNLI across three of the five source datasets. Conversely, fine-tuning on translated NLI improves performance on \textsc{HEntailment} for weaker LLMs, but degrades performance for the strongest models, with losses concentrated on neutral examples.
This asymmetric transfer shows that translated and native NLI data are not reliable substitutes for one another. Our approach reuses existing task annotations and is designed to generalize beyond Hebrew, offering a practical path toward native-language NLI benchmarks without translating English evaluation suites.
\end{abstract}

\section{Introduction}
\label{sec:introduction}

The task of {\em Natural Language Inference} (NLI), determining whether a hypothesis is entailed by, contradicted by, or neutral with respect to an input premise \citep{dagan2005pascal}, has long served as a core task of language understanding \citep{dagan2013recognizing, madaan-etal-2025-lost}. English corpora such as SNLI \citep{bowman-etal-2015-large} and MultiNLI \citep{williams-etal-2018-broad} made it a standard instrument for evaluating \citep{wang-etal-2018-glue, brown2020language} and fine-tuning \citep{wei2022finetuned, sanh2022multitask} large language models (LLMs).
However, many lower-resource languages lack comparable native resources \citep{joshi-etal-2020-state}, as expert annotation is costly. Their NLI competence in these languages therefore cannot be measured on native text.

The common workaround is translation of English benchmarks, as in XNLI \citep{conneau2018xnli}, in order to expand the  multilingual reach of the NLI task.
Translation makes cross-lingual comparison possible, but it is sub-optimal, as translation, especially automatic, may yield {\em translationese} \citep{koppel2011translationese}: text that is systematically simpler, more literal, and structurally closer to the source language than authentic target-language text \citep{toral2019post, vanmassenhove-etal-2021-machine}.

To overcome this resource gap and obtain NLI resources that preserve the native language properties, we propose to `translate' tasks that contain native data in that language into NLI. Concretely, we propose to repurpose native annotations that already exist in a range of tasks --- a question with its gold answer, a summary and its source, a paraphrase pair, etc. ---  into NLI pairs.
By ensuring the text never leaves its source language, native linguistic and cultural content is preserved by design, without requiring a new round of expert annotation.

To operationalize this approach, we design a repurposing framework with three components. (1) Entailment: rule-based alignment maps each source annotation onto pairs whose label follows deterministically from the original task; an LLM performs surface realization. (2) Contradiction: targeted heuristics edit the entailed hypothesis. An LLM filters these edits, discarding those that are ungrammatical or do not contradict the premise. (3) Neutral: counterparts are selected by similarity ranking over sentence embeddings.

Using Hebrew as the  case study, we introduce \textsc{HEntailment}, to our knowledge the first Hebrew NLI benchmark based on native Hebrew data, consisting of 173{,}460 premise--hypothesis pairs (160{,}847 training and development pairs and 12{,}613 test pairs) repurposed from five native datasets: the QA corpora HeQ \citep{cohen-etal-2025-hebrewqa} and Hebrew QASem \citep{davidov-etal-2026-effective}, the summarization corpora HeSum \citep{paz-argaman-etal-2024-hesum} and HebSummaries \citep{hebarabnlp2024hebsummaries}, and Hebrew-Paraphrase \citep{hebarabnlp2024paraphrase}.
The labels hold up under human review. Two native Hebrew speakers independently labeled a stratified sample of 315 items and agreed with each other at $\kappa = 0.763$. Against the pipeline's labels, the annotators reach up to 0.89 macro-F1. The benchmark also leaves headroom: two frontier LLMs prompted zero-shot reach only 0.788 and 0.728 macro-F1 against its labels (Table~\ref{tab:zeroshot}), below both annotators.

We use \textsc{HEntailment} to assess transfer between native and translated NLI in Hebrew, using the two existing English-to-Hebrew translated benchmarks, HebNLI \citep{nnlp2024hebnli} and LCHAIM \citep{malul-etal-2025-lchaim}. Transfer from \textsc{HEntailment} to the translated benchmarks depends on the source task and on the architecture. Fine-tuning a decoder on HeQ, QASem or Hebrew-Paraphrase repurposed NLI pairs raises its HebNLI score by 0.06 to 0.08 on average. On LCHAIM no source helps by more than 0.012. After fine-tuning on any single source, every decoder outperforms every encoder on HebNLI. Transfer in the other direction is also limited. The direction of the change depends on the model (Figure~\ref{fig:transfer}): averaged over the five test sets, the three decoders with the lowest zero-shot scores gain 0.01 to 0.19, while the three strongest lose 0.26 to 0.29, mostly on neutral pairs. We hypothesize that these differences are partly explained by characteristics specific to translated benchmarks, including less authentic structure and disparate knowledge and culture. This asymmetry shows that translated and native NLI data are not reliable substitutes for one another.

These findings call for native NLI evaluation, and it need not wait for expert annotation. Our approach\footnote{Code, data and models: \url{https://github.com/Ronke21/HEntailment-Hebrew-NLI-Repurposing}} turns the annotated tasks a language already has into NLI, using only native task annotations, a sentence encoder and an LLM fluent in the language, a combination many lower-resource and mid-resource languages already have. It is designed to generalize beyond Hebrew without translating English evaluation suites, and we advocate complementing translated benchmarks with benchmarks derived from native-language tasks.

\section{The Vicious Cycle of Translation in Benchmark Construction}
\label{sec:cycle}

Low-resource languages are characterized by a scarcity of annotated task data and under-representation in pre-training corpora, a status largely independent of their global speaker population \citep{joshi-etal-2020-state}.
When such a language lacks a resource such as NLI, the cheapest solution is to translate an English one; XNLI \citep{conneau2018xnli} set the pattern, and for many languages the only NLI data is translated. However, translated text is not neutral: it may exhibit translationese, being systematically simpler, more literal and structurally closer to the source language than authentic production \citep{koppel2011translationese}. Machine translation and post-editing amplify this effect \citep{toral2019post, vanmassenhove-etal-2021-machine}, sometimes enough to distort cross-lingual evaluation \citep{artetxe-etal-2020-translation}.

For a language such as Hebrew \citep{tsarfaty2019whats} the mismatch is not cosmetic. Translation from English may preserve English constituent order and spell out material that Hebrew encodes morphologically, so translated Hebrew under-represents exactly the phenomena that pose challenges for LLMs processing Hebrew: rich inflection, construct-state phrases, and orthographic ambiguity in unvocalized writing.

The problem compounds because translated data sits at both ends of the pipeline: multilingual models fine-tuned on translated corpora are evaluated on translated benchmarks, so a model that has learned translationese is certified by data exhibiting the same regularities, and evaluation cannot separate competence in the language from robustness to translation.
LLMs also default to US-centric answers to locale-ambiguous questions, even when asked in other languages \citep{mor-lan-etal-2026-location}. Translated benchmarks alone therefore cannot measure cultural adaptation: they embed the source locale's context and can mask English-anchored defaults.

Hebrew NLI is a clear instance. Both available resources are translations: HebNLI \citep{nnlp2024hebnli} renders MultiNLI into Hebrew, and LCHAIM \citep{malul-etal-2025-lchaim} translates and validates the English ConTRoL corpus. To our knowledge no natively annotated Hebrew NLI dataset exists, so a Hebrew model's NLI competence is currently measurable only on text not originally authored in Hebrew.

\section{Native Inference via Repurposing}
\label{sec:proposal}

To obtain native NLI benchmarks, we propose to repurpose task annotations available in the target language (e.g., QA or summarization) instead of translating English NLI resources. A language may lack annotated NLI data while already possessing high-quality datasets for related semantic tasks. Hebrew, for example, has native data annotated for question answering, summarization, and paraphrases, whose annotations naturally encode inferential relations: a question with its gold answer forms a proposition supported by its passage, a faithful summary is entailed by its source, and paraphrases in fact stand in mutual entailment.

Repurposing task annotations into NLI is not itself new. English benchmarks such as QNLI were derived from SQuAD \citep{wang-etal-2018-glue}, question answering has been reformulated as inference \citep{demszky2018transforming}, and similar reformulations have been applied to relation extraction \citep{sainz-etal-2021-label}. Our contribution is to argue that repurposing is a general method for constructing native NLI benchmarks in lower-resource languages, and to test it against the translated alternative. So, whereas repurposing in English primarily targets increased benchmark scale,  and reduced costs, in our setting it aims to preserve the linguistic phenomena that translation flattens.

The method we propose takes an annotated dataset as input  and emits three-way NLI pairs, by three separate mechanisms. The \emph{premise} is always a span of authentic source text. \textbf{Entailment} is read off the annotation itself, with the label being derived: whatever relation the source task certifies (a question with its gold answer, a summary with its article, a paraphrase with its counterpart) is intended to follow directly from the original annotation. A language model is thus used only to render that relation as a fluent standalone sentence (e.g. combine question+answer). \textbf{Contradiction} perturbs an entailed hypothesis with a rule that changes its truth conditions, so this label is rule-derived. An LLM judge discards candidates that are ungrammatical or that it does not read as contradictions. \textbf{Neutral} pairs a premise with a hypothesis drawn from a \emph{different} document, chosen by sentence-embedding similarity so the pair is topically plausible but not entailed.

Having introduced the repurposing framework, we examine whether the resulting benchmark captures NLI abilities. Because no natively annotated Hebrew NLI benchmark exists, a direct comparison against native gold data is impossible. Instead, we evaluate the proposed benchmark through three complementary questions: (1) is the repurposed data learnable; (2) does competence acquired from it transfer to independent translated NLI benchmarks; and (3) does competence acquired from translated NLI transfer into the repurposed benchmark? Together, these experiments assess both the practical utility of the benchmark and the broader implications of repurposing native task annotations for NLI evaluation.

\section{HEntailment: The Hebrew NLI Benchmark}
\label{sec:benchmark}

Using the proposed procedure (\S\ref{sec:proposal}) we deliver \textsc{HEntailment}, which  is, to our knowledge, the first Hebrew NLI resource constructed entirely from native Hebrew sources, rather than translated English benchmarks. Table~\ref{tab:hentailment} provides the basic statistics of the released corpus; per-source counts are in Appendix~\ref{app:data}.

\begin{table}[t]
\centering\small
\setlength{\tabcolsep}{4pt}
\begin{tabular}{lrrrr}
\toprule
& \textbf{Ent.} & \textbf{Con.} & \textbf{Neu.} & \textbf{Total} \\
\midrule
Train & 65{,}098 & 25{,}253 & 58{,}844 & 149{,}195 \\
Dev   &  4{,}912 &  2{,}546 &  4{,}194 &  11{,}652 \\
Test  &  5{,}334 &  2{,}668 &  4{,}611 &  12{,}613 \\
\midrule
\textbf{Total} & \textbf{75{,}344} & \textbf{30{,}467} & \textbf{67{,}649} & \textbf{173{,}460} \\
\bottomrule
\end{tabular}
\caption{\textsc{HEntailment} statistics by split and label. No premise is shared between train and test.}
\label{tab:hentailment}
\end{table}

\textsc{HEntailment} is repurposed from five Hebrew datasets spanning question answering, summarization and paraphrase; Table~\ref{tab:sources} gives their sizes and provenance (tables abbreviate HebSummaries as HebSumm. and Hebrew-Paraphrase as HebPara.). All five source datasets are in Hebrew, so every premise is native text; hypotheses come from the source annotations, with some rewritten by an LLM or edited by rules.

\begin{table}[t]
\centering\small
\setlength{\tabcolsep}{4pt}
\begin{tabular}{llrrc}
\toprule
& & \multicolumn{2}{c}{\textbf{converted}} & \\
\cmidrule(lr){3-4}
\textbf{Source} & \textbf{Task} & \textbf{Train} & \textbf{Test} & \textbf{Native} \\
\midrule
HeQ                & QA         & 47{,}505 & 2{,}639 & yes \\
QASem        & QA-SRL     & 42{,}852 & 1{,}722 & yes \\
HeSum              & summ.\     & 19{,}072 & 2{,}437 & yes \\
HebSumm.\          & summ.\     & 10{,}111 & 2{,}102 & yes \\
HebPara.\       & paraphrase & 29{,}655 & 3{,}713 & yes \\
\midrule
\multicolumn{5}{l}{\emph{existing Hebrew NLI, not repurposed}} \\
HebNLI             & NLI        & 300{,}067$^{b}$ & 884 & no$^{a}$ \\
LCHAIM             & NLI        & 6{,}719$^{c}$ & 805 & no$^{a}$ \\
\bottomrule
\end{tabular}
\caption{Source datasets and the pairs they convert to.  $^{a}$~Translated from English. $^{b}$~The transfer experiments train on a 15{,}000-example subsample (HebNLI-15k). $^{c}$~Used for training only as LCHAIM-max (5{,}838 examples).}
\label{tab:sources}
\end{table}

\paragraph{Corpus creation.} We apply the method of \S\ref{sec:proposal} with one mechanism per label (Table~\ref{tab:labelsources}).

\textbf{Entailment.} The entailed hypothesis comes from the source annotation itself. For the two QA sources, an LLM (\texttt{gemini-3.1-flash-lite}) turns a question and its gold answer into a standalone declarative sentence; the summarization and paraphrase sources use their own text, which is already aligned.

\textbf{Contradiction.} Hebrew transformation heuristics edit the entailed hypothesis. Typical rules swap a person, organization or place for another entity of the same type, replace a number with a plausible one of the same format, or flip an adjective to its antonym (Figure~\ref{fig:twin}); Appendix~\ref{app:strategies} lists every rule. We exclude six rules, such as verb negation, because a hypothesis-only classifier can distinguish their contradictions from entailments. An LLM judge (gemma-4-31B-it) keeps a candidate only if the edited sentence is grammatical and the pair is a contradiction, preferring edits to a detail the premise states. Rules turn 40\% of the 75{,}344 entailment hypotheses into a valid contradiction, so contradiction is the minority class (17.6\% of the corpus).

\textbf{Neutral.} Each premise is paired with a hypothesis written for a different document. We use NeoDictaBERT \citep{shmidman2025neodictabertpushingfrontierbert} embeddings to rank the candidates, and the partner is drawn from the most similar ones below a cosine ceiling of 0.7. Sampling a distant sentence instead leaves the class recoverable from topic alone (Appendix~\ref{app:revised}).

\begin{table*}[t]
\centering\small
\setlength{\tabcolsep}{5pt}
\renewcommand{\arraystretch}{1.15}
\begin{tabular}{@{}>{\raggedright\arraybackslash}p{2.2cm}>{\raggedright\arraybackslash}p{4.3cm}>{\raggedright\arraybackslash}p{4.3cm}>{\raggedright\arraybackslash}p{3.9cm}@{}}
\toprule
& \textbf{QA} (HeQ, QASem) & \textbf{Summarization} (HeSum, HebSumm.) & \textbf{Paraphrase} (HebPara.) \\
\midrule
\textbf{Premise} & passage (HeQ) or sentence (QASem) & article or document & one sentence of the pair \\
\addlinespace
\textbf{Entailment} & question + gold answer, rewritten as a statement by an LLM & the human-written summary & the other sentence (both directions) \\
\addlinespace
\textbf{Contradiction} & \multicolumn{3}{>{\raggedright\arraybackslash}p{13.0cm}@{}}{a rule edit of the entailment hypothesis (entity swap, number shift, antonym, \ldots), kept only if an LLM judge finds it grammatical and contradictory} \\
\addlinespace
\textbf{Neutral} & \multicolumn{3}{>{\raggedright\arraybackslash}p{13.0cm}@{}}{a hypothesis from another document, the most similar one under NeoDictaBERT embeddings (cosine below 0.7)} \\
\bottomrule
\end{tabular}
\caption{How each label is built for each source family. Contradiction and neutral are built the same way for all five sources.}
\label{tab:labelsources}
\end{table*}

\paragraph{\textsc{HEntailment} Validation.}
\label{sec:validation}
We validate the benchmark four ways: against human annotators, against a hypothesis-only probe, against frontier LLMs prompted zero-shot, and against an independent LLM judge on the whole test set.

\textbf{Human validation.} Two annotators, both native Hebrew speakers, independently relabeled a stratified sample from the test split, drawn per (source $\times$ label) cell under a fixed seed. Each saw only the premise and the hypothesis, never the pipeline's label. The scheme has four options: the three NLI labels, or a fourth marking the hypothesis incoherent, so that an unparsable item is set aside rather than forced into a label. Appendix \ref{app:guidelines} reproduces the guidelines.

We sampled 315 test items (63 per source, 105 per label). The annotators agree with each other at Cohen's $\kappa = 0.763$, and the annotators reach 0.822 and 0.890 macro-F1 against the automatic labels. To check the contradictions in the released corpus, a second judge from another model family that took no part in construction (Qwen3.6-27B) agrees with the pipeline on 89\% of 2{,}000 sampled contradictions. Appendix~\ref{app:human} gives the full breakdown.

\textbf{Hypothesis-only NLI validation.} Following \citet{poliak-etal-2018-hypothesis}, classifiers are trained on hypotheses alone, with premises withheld, and evaluated on a class-balanced sample of 708 items (236 per label) per dataset; the \textsc{HEntailment} sample is additionally balanced across its five sources. Every condition trains on the same class-balanced 1{,}900 examples per label, the largest balanced training subset all datasets support, so differences reflect dataset artifacts rather than training-set size. We report two probe capacities: TF-IDF unigrams with logistic regression, and multilingual-E5 fine-tuned under the hyper-parameters of \citet{proebsting-poliak-2024-hypothesis}.

\begin{table}[t]
\centering\small
\begin{tabular}{lcc}
\toprule
\textbf{Dataset} & \textbf{TF-IDF} & \textbf{mE5} \\
\midrule
\textsc{HEntailment} & 0.407 & 0.424 \\
\midrule
\emph{HebNLI} & \emph{0.470} & \emph{0.578} \\
\emph{LCHAIM} & \emph{0.435} & \emph{0.466} \\
\bottomrule
\end{tabular}
\caption{Hypothesis-only accuracy on 708 class-balanced items per dataset (chance 0.333, 95\% interval $\pm 0.036$).}
\label{tab:hyponly}
\end{table}

Under both probes (Table~\ref{tab:hyponly}), \textsc{HEntailment} is less recoverable from the hypothesis alone than HebNLI (TF-IDF 0.407 against 0.470, mE5 0.424 against 0.578), by more than the $\pm 0.036$ interval, and neither reaches the 0.72 \citet{proebsting-poliak-2024-hypothesis} report for \emph{human-written} SNLI. HebNLI is itself substantially solvable without its premises, in line with \citet{poliak-etal-2018-hypothesis} across ten English corpora. On the full corpus, with folds grouped by premise, a hypothesis-only classifier does worse than predicting the majority class on four of five sources.

\textbf{Zero-shot validation.} We prompt two frontier LLMs zero-shot over the Hebrew pair, on the same 315 items the annotators scored, and evaluate them against the pipeline's labels (Table~\ref{tab:zeroshot}). The benchmark leaves headroom: the stronger model reaches 0.788 macro-F1 and the weaker 0.728, both below the two annotators (0.822 and 0.890).

\begin{table}[t]
\centering\small
\begin{tabular}{lccc}
\toprule
\textbf{Model} & \textbf{Macro-F1} & \textbf{Acc.} & \textbf{$\kappa$} \\
\midrule
\emph{human A1}       & \emph{0.822} & \emph{0.829} & \emph{0.742} \\
\emph{human A2}       & \emph{0.890} & \emph{0.891} & \emph{0.836} \\
\midrule
gemini-3.5-flash-lite & 0.788 & 0.799 & 0.697 \\
gpt-5.4-mini          & 0.728 & 0.737 & 0.605 \\
\bottomrule
\end{tabular}
\caption{Judges against \textsc{HEntailment}'s labels on the same 315 items: two annotators (italic) and two zero-shot LLMs. Breakdown in Appendix~\ref{app:zeroshot}.}
\label{tab:zeroshot}
\end{table}

Every judge finds \texttt{contradiction} hardest, with per-class agreement from 0.457 for gpt-5.4-mini to 0.812 for annotator 2. Appendix~\ref{app:zeroshot} gives the per-source and per-label breakdown and each judge's agreement with the human reference.

\textbf{Verified test subset.} Qwen3.6-27B, the second judge above, also labels every test item zero-shot with the construction judge's prompt. It confirms 11{,}071 of the 12{,}613 test items (87.8\%). Re-scored on these items, the mean over the five \textsc{HEntailment} test sets rises by 0.014 and the model ranking is preserved ($\rho = 0.998$ over 94 model and condition cells). Because Qwen3.6-27B is also an evaluated model, we use the confirmed items, which we release as a secondary test set, only as a check on the full test set (Appendix~\ref{app:robust}).

\section{Native vs.\ Translated Transfer Experiments}
\label{sec:setup}

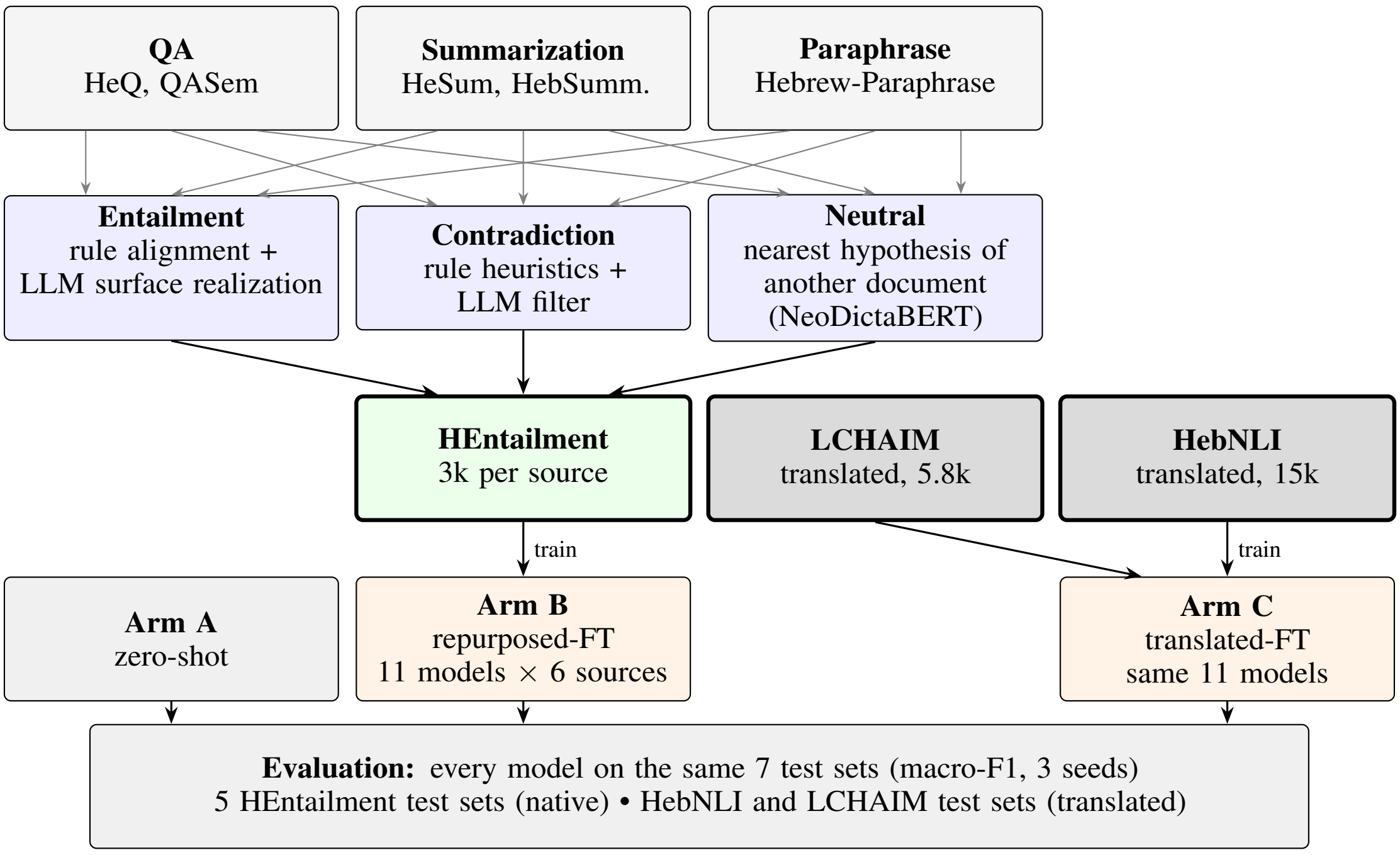
\begin{figure*}[t]
\centering
\footnotesize
\begin{tikzpicture}[
  box/.style   ={draw, rounded corners=2pt, align=center, inner sep=3pt,
                 font=\footnotesize, text width=33mm, minimum height=13mm},
  src/.style   ={box, fill=black!4},
  proc/.style  ={box, fill=blue!7},
  data/.style  ={box, fill=green!8, very thick},
  ref/.style   ={box, fill=black!14, very thick},
  exp/.style   ={box, fill=orange!10},
  evalb/.style ={box, fill=black!6},
  ar/.style    ={-{Stealth[length=1.8mm]}, semithick},
  lat/.style   ={-{Stealth[length=1.4mm]}, thin, black!50},
]
\def\cA{0} \def\cB{3.7} \def\cC{7.4} \def\cD{11.1}
\node[src] (qa)  at (\cA,0) {\textbf{QA}\\HeQ, QASem};
\node[src] (sum) at (\cB,0) {\textbf{Summarization}\\HeSum, HebSumm.};
\node[src] (par) at (\cC,0) {\textbf{Paraphrase}\\Hebrew-Paraphrase};
\node[proc] (ent) at (\cA,-2.1) {\textbf{Entailment}\\rule alignment +\\LLM surface realization\\[1pt]\mbox{}};
\node[proc] (con) at (\cB,-2.1) {\textbf{Contradiction}\\rule heuristics +\\LLM filter};
\node[proc] (neu) at (\cC,-2.1) {\textbf{Neutral}\\nearest hypothesis of\\another document\\(NeoDictaBERT)};
\draw[lat] ([xshift=-9mm]qa.south)  -- ([xshift=-9mm]ent.north);
\draw[lat] (qa.south)               -- ([xshift=-9mm]con.north);
\draw[lat] ([xshift=9mm]qa.south)   -- ([xshift=-9mm]neu.north);
\draw[lat] ([xshift=-9mm]sum.south) -- (ent.north);
\draw[lat] (sum.south)              -- (con.north);
\draw[lat] ([xshift=9mm]sum.south)  -- (neu.north);
\draw[lat] ([xshift=-9mm]par.south) -- ([xshift=9mm]ent.north);
\draw[lat] (par.south)              -- ([xshift=9mm]con.north);
\draw[lat] ([xshift=9mm]par.south)  -- ([xshift=9mm]neu.north);
\node[data] (hent) at (\cB,-4.1)
  {\textbf{HEntailment}\\3k per source};
\node[ref] (heb) at (\cD,-4.1)
  {\textbf{HebNLI}\\ translated, 15k};
\node[ref] (lca) at (\cC,-4.1)
  {\textbf{LCHAIM}\\ translated, 5.8k};
\draw[ar] (ent.south) -- ([xshift=-9mm]hent.north);
\draw[ar] (con.south) -- (hent.north);
\draw[ar] (neu.south) -- ([xshift=9mm]hent.north);
\node[evalb] (zs)  at (\cA,-6.0) {\textbf{Arm A}\\zero-shot};
\node[exp]   (rft) at (\cB,-6.0) {\textbf{Arm B}\\repurposed-FT\\11 models $\times$ 6 sources};
\node[exp]   (nft) at (\cD,-6.0) {\textbf{Arm C}\\translated-FT\\same 11 models};
\draw[ar] (hent.south) -- node[right,font=\scriptsize] {train} (rft.north);
\draw[ar] (heb.south)  -- node[right,font=\scriptsize] {train} (nft.north);
\draw[ar] (lca.south)  -- ([xshift=-9mm]nft.north);
\node[evalb, text width=126mm] (ev) at (5.55,-7.55)
  {\textbf{Evaluation:} every model on the same 7 test sets (macro-F1, 3 seeds)\\
   5 HEntailment test sets (native) \textbullet\ HebNLI and LCHAIM test sets (translated)};
\draw[ar] (zs.south)  -- (zs.south  |- ev.north);
\draw[ar] (rft.south) -- (rft.south |- ev.north);
\draw[ar] (nft.south) -- (nft.south |- ev.north);
\end{tikzpicture}
\caption{Experimental design: repurposing and bidirectional transfer evaluation.}
\label{fig:design}
\end{figure*}

\paragraph{Goal.} Using \textsc{HEntailment}, we ask whether NLI competence transfers between native repurposed data and translated Hebrew NLI, and whether it does so symmetrically.

\paragraph{Scenarios.} Figure~\ref{fig:design} summarizes the design. \emph{Native~$\rightarrow$~translated} fine-tunes each model on one repurposed source, or on the pooled mixture \textsc{all}, and scores it on two translated Hebrew NLI benchmarks, \textbf{HebNLI} \citep{nnlp2024hebnli} and \textbf{LCHAIM} \citep{malul-etal-2025-lchaim} (Table~\ref{tab:translated}); each is used for training only in its own in-task control. \emph{Translated~$\rightarrow$~native} takes the HebNLI-fine-tuned checkpoints and scores them on the five held-out \textsc{HEntailment} test sets (\S\ref{sec:benchmark}). Every model is evaluated on all seven test sets, so the two directions share one harness.

\begin{table}[t]
\centering\small
\setlength{\tabcolsep}{2.4pt}
\begin{tabular}{llrrcr}
\toprule
\textbf{Dataset} & \textbf{From} & \textbf{Train} & \textbf{Test} & \textbf{E/C/N} & \textbf{Subset} \\
\midrule
HebNLI & MultiNLI & 300{,}067 & 884 & 307/288/289 & 15{,}000 \\
LCHAIM & ConTRoL  & 6{,}719   & 805 & 327/242/236 & 5{,}838 \\
\bottomrule
\end{tabular}
\caption{The two translated Hebrew NLI benchmarks and the English corpora they translate. \emph{E/C/N}: test labels. \emph{Subset}: the class-balanced HebNLI-15k and LCHAIM-max training subsets, used only as in-task controls.}
\label{tab:translated}
\end{table}

\paragraph{Models.}
We fine-tune eleven models: five encoders (AlephBERT, NeoDictaBERT, XLM-R large and XL, multilingual-E5-large) and six decoders (DictaLM-3.0 1.7B-Instruct and 24B-Base, Hebatron \citep{hebatron2026}, gemma-4-12B-it and 31B-it \citep{gemma2026gemma4}, Qwen3.6-27B \citep{qwen2026qwen36}). The two frontier LLMs of \S\ref{sec:benchmark} are validators only and are not fine-tuned. Tables omit the \texttt{-it} and \texttt{-Instruct} suffixes of model names. Encoders are fully fine-tuned with a 3-way classification head; decoders with LoRA ($r{=}16$, $\alpha{=}32$, all-linear) and predict a Hebrew label word by greedy decoding, followed by label normalization. Prompts, targets and label normalization are identical across conditions. Every hyperparameter and the hardware are in Appendix~\ref{app:setup}.

\paragraph{Training scenarios.}
\label{sec:setup-hebnli}
Each model is trained under eight conditions: each of the five sources alone (1{,}000 examples per label), the five pooled (\textsc{all}), HebNLI-15k (15{,}000 examples, the same budget as the pool) and LCHAIM-max (5{,}838 examples, the second in-task control). Every condition runs with three seeds (42, 43, 44), and we report means and seed standard deviations: 270 runs including the decoders' zero-shot scores.

\paragraph{Metrics.} Macro-F1 throughout. Transfer is reported as a delta against \emph{the same model's own zero-shot score}, so each model is its own control: absolute fine-tuned scores conflate transfer with pretrained ability.

\section{Transfer Results and Analysis}
\label{sec:results}

\subsection{Native to Translated}
\label{sec:res}

\paragraph{In-task fine-tuning sets the ceiling.} We score models fine-tuned on \textsc{HEntailment} on the two translated Hebrew NLI benchmarks. As a ceiling, fine-tuning on 15{,}000 HebNLI examples gives the eleven models an average of 0.808 macro-F1 on HebNLI (decoders 0.898, encoders 0.700).

Table~\ref{tab:transfer} reports, per training source, how often fine-tuning improves a decoder over its own zero-shot score on the two held-out Hebrew NLI benchmarks.

\begin{table}[t]
\centering\small
\setlength{\tabcolsep}{4pt}
\resizebox{\columnwidth}{!}{%
\begin{tabular}{lcccc}
\toprule
\textbf{Training source} & \textbf{Impr.} & \textbf{Mean $\Delta$} & \textbf{$\Delta$ HebNLI} & \textbf{$\Delta$ LCHAIM} \\
\midrule
HeQ               & 7/12 & $+0.041$ & $+0.071$ & $+0.012$ \\
QASem             & 7/12 & $+0.012$ & $+0.061$ & $-0.036$ \\
HeSum             & 1/12 & $-0.080$ & $-0.108$ & $-0.051$ \\
HebSumm.      & 5/12 & $-0.021$ & $-0.023$ & $-0.018$ \\
HebPara. & 8/12 & $+0.033$ & $+0.075$ & $-0.010$ \\
\textsc{all} (pooled) & 5/12 & $-0.018$ & $+0.011$ & $-0.047$ \\
\midrule
\emph{HebNLI-15k}  & 11/12 & $+0.138$ & $+0.203$ & $+0.074$ \\
\emph{LCHAIM-max}  & 9/12 & $+0.070$ & $-0.034$ & $+0.173$ \\
\bottomrule
\end{tabular}}
\caption{Transfer to held-out Hebrew NLI: change against each decoder's own zero-shot score, mean of three seeds (six decoders $\times$ two benchmarks per row). Italic rows are in-task controls.}
\label{tab:transfer}
\end{table}

\begin{table*}[t]
\centering\small
\setlength{\tabcolsep}{6pt}
\begin{tabular}{lccccccc}
\toprule
\textbf{Train} & \textbf{HeQ} & \textbf{QASem} & \textbf{HeSum} & \textbf{HebSumm.} & \textbf{HebPara.} & \textbf{\emph{HebNLI}} & \textbf{\emph{LCHAIM}} \\
\midrule
zero-shot & \cellcolor{blue!46}.46 & \cellcolor{blue!44}.44 & \cellcolor{blue!36}.36 & \cellcolor{blue!37}.37 & \cellcolor{blue!25}.25 & \cellcolor{blue!51}.51 & \cellcolor{blue!27}.27 \\
HeQ & \cellcolor{blue!97}\color{white}\textbf{.97} & \cellcolor{blue!85}\color{white}.85 & \cellcolor{blue!72}\color{white}.72 & \cellcolor{blue!72}\color{white}.72 & \cellcolor{blue!86}\color{white}.86 & \cellcolor{blue!67}\color{white}.67 & \cellcolor{blue!41}.41 \\
QASem & \cellcolor{blue!93}\color{white}.93 & \cellcolor{blue!93}\color{white}\textbf{.93} & \cellcolor{blue!60}\color{white}.60 & \cellcolor{blue!51}.51 & \cellcolor{blue!83}\color{white}.83 & \cellcolor{blue!67}\color{white}.67 & \cellcolor{blue!38}.38 \\
HeSum & \cellcolor{blue!95}\color{white}.95 & \cellcolor{blue!73}\color{white}.73 & \cellcolor{blue!92}\color{white}\textbf{.92} & \cellcolor{blue!79}\color{white}.79 & \cellcolor{blue!85}\color{white}.85 & \cellcolor{blue!46}.46 & \cellcolor{blue!32}.32 \\
HebSumm. & \cellcolor{blue!94}\color{white}.94 & \cellcolor{blue!68}\color{white}.68 & \cellcolor{blue!84}\color{white}.84 & \cellcolor{blue!93}\color{white}\textbf{.93} & \cellcolor{blue!91}\color{white}.91 & \cellcolor{blue!53}.53 & \cellcolor{blue!36}.36 \\
HebPara. & \cellcolor{blue!80}\color{white}.80 & \cellcolor{blue!83}\color{white}.83 & \cellcolor{blue!64}\color{white}.64 & \cellcolor{blue!66}\color{white}.66 & \cellcolor{blue!97}\color{white}\textbf{.97} & \cellcolor{blue!68}\color{white}.68 & \cellcolor{blue!35}.35 \\
\midrule
\textsc{all} & \cellcolor{blue!98}\color{white}.98 & \cellcolor{blue!95}\color{white}.95 & \cellcolor{blue!94}\color{white}.94 & \cellcolor{blue!94}\color{white}.94 & \cellcolor{blue!98}\color{white}.98 & \cellcolor{blue!63}\color{white}.63 & \cellcolor{blue!40}.40 \\
\bottomrule
\end{tabular}
\caption{\textbf{DictaLM-3.0-1.7B-Instruct} fine-tuned on each source (macro-F1, mean of three seeds; darker is higher), the decoder with the largest transfer gain; other models' in-domain, HebNLI and LCHAIM scores in Appendix~\ref{app:heatmaps}. \textbf{Bold}: in-domain. Italic columns: translated NLI.}
\label{tab:heatbest}
\end{table*}

\paragraph{Native training helps for three of five sources.} Fine-tuning on \textsc{HEntailment} helps decoders on HebNLI when the source is HeQ, QASem or Hebrew-Paraphrase ($+0.06$ to $+0.08$) and hurts when it is HeSum ($-0.11$). Pooling the five sources does not add up ($+0.011$ on HebNLI). On LCHAIM only HeQ is positive on average, by 0.012. In-task supervision is the control: HebNLI-15k raises decoders by 0.203 on HebNLI and 0.074 on LCHAIM, positive in 11 of 12 cells.

\paragraph{The gain depends on the model.} The HeQ effect is an example: the Hebrew-native decoders gain 0.09 to 0.15 on HebNLI, and the gemma models change by $-0.027$ and $-0.005$ (Appendix~\ref{app:permodel}).

\subsection{Translated to Native}
\label{sec:reverse}

\paragraph{Translated training hurts the strongest models.} Models fine-tuned on HebNLI-15k are scored on the five held-out \textsc{HEntailment} test sets (Table~\ref{tab:reverse}). Fine-tuning on translated NLI helps decoders on \textsc{HEntailment} only on QASem ($+0.063$). On the other four sources it lowers the score by 0.10 to 0.20 on average, and most for the strongest zero-shot models: gemma-4-31B loses 0.05 to 0.46 and Qwen3.6-27B 0.05 to 0.37.

\paragraph{The weakest decoder gains.} DictaLM-3.0-1.7B gains on four of five sources ($+0.08$ to $+0.38$); zero-shot it labels 77\% of items neutral.

\begin{table}[t]
\centering\footnotesize
\setlength{\tabcolsep}{4pt}
\resizebox{\columnwidth}{!}{%
\begin{tabular}{lcccccc}
\toprule
\textbf{Decoder} & \textbf{HeQ} & \textbf{QASem} & \textbf{HeSum} & \textbf{HebSumm.} & \textbf{HebPara.} & \textbf{N$\to$C} \\
\midrule
DictaLM-3.0-1.7B & $+.20$ & $+.28$ & $+.08$ & $-.01$ & $+.38$ & 6$\to$30 \\
DictaLM-3.0-24B & $+.02$ & $+.14$ & $+.18$ & $-.02$ & $+.06$ & 14$\to$30 \\
Hebatron         & $-.07$ & $+.13$ & $+.06$ & $+.03$ & $-.09$ & 19$\to$40 \\
gemma-4-12B      & $-.28$ & $-.08$ & $-.31$ & $-.39$ & $-.26$ & 2$\to$75 \\
Qwen3.6-27B      & $-.29$ & $-.05$ & $-.36$ & $-.37$ & $-.36$ & 3$\to$67 \\
gemma-4-31B      & $-.22$ & $-.05$ & $-.42$ & $-.46$ & $-.30$ & 3$\to$59 \\
\midrule
\emph{Mean}      & $-.11$ & $+.06$ & $-.13$ & $-.20$ & $-.10$ & \\
\bottomrule
\end{tabular}}
\caption{Translated to native: change in macro-F1 against zero-shot for each decoder fine-tuned on HebNLI-15k, on the five \textsc{HEntailment} test sets (mean of three seeds; decoders ordered by zero-shot score). \emph{N$\to$C}: percentage of gold-neutral items labeled contradiction, zero-shot $\to$ fine-tuned.}
\label{tab:reverse}
\end{table}

\subsection{Analysis}

\paragraph{Decoders learn every source.} In-domain macro-F1 is 0.90 to 0.99 in all but one cell (gemma-4-12B on HebSummaries, $0.833 \pm 0.091$), and the pooled mixture reaches 0.94 to 0.98. Encoders learn less reliably: the pooled mixture averages 0.76 (NeoDictaBERT) to 0.91 (XLM-R-XL) (Appendix~\ref{app:permodel}). Learnability and transfer come apart by architecture (Tables~\ref{tab:indomain}--\ref{tab:permodel_lcahim}).

\paragraph{Only decoders transfer.} After fine-tuning on a single \textsc{HEntailment} source, every decoder scores higher on HebNLI than every encoder: decoders 0.600 to 0.788, encoders 0.279 to 0.411 (mean over the five sources; chance 0.333). With the model as the unit, a one-sided exact Mann-Whitney test gives $U = 30$ of 30, $p = 0.0022$. On LCHAIM no encoder learns from any source (0.20 to 0.34).

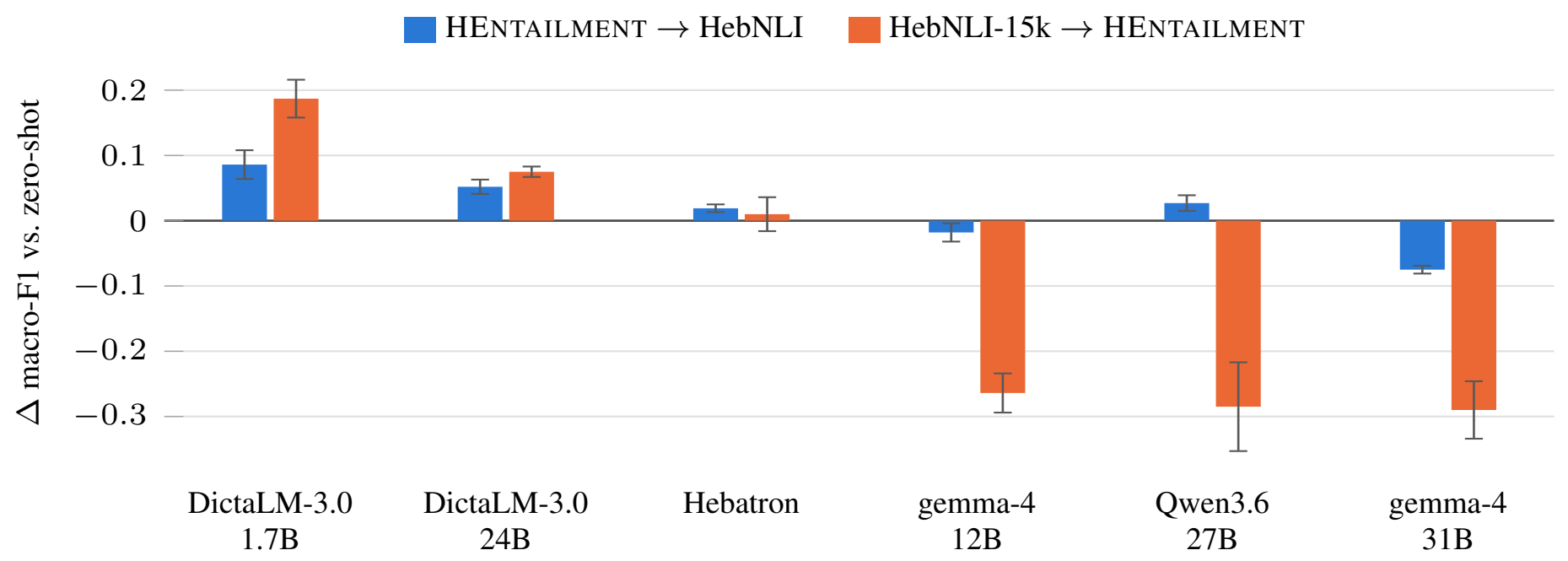
\begin{figure*}[t]
\centering
\definecolor{hentblue}{HTML}{2A78D6}\definecolor{hentorange}{HTML}{EB6834}
\begin{tikzpicture}
\begin{axis}[
  width=0.78\textwidth, height=4.6cm,
  ybar=1.5pt, bar width=10pt,
  ymin=-0.36, ymax=0.23,
  ylabel={$\Delta$ macro-F1 vs.\ zero-shot}, ylabel style={font=\scriptsize},
  symbolic x coords={d17,d24,heb,g12,q27,g31},
  xtick=data,
  xticklabels={DictaLM-3.0\\1.7B, DictaLM-3.0\\24B, Hebatron, gemma-4\\12B, Qwen3.6\\27B, gemma-4\\31B},
  xticklabel style={align=center, font=\scriptsize},
  yticklabel style={font=\scriptsize, /pgf/number format/fixed, /pgf/number format/precision=1},
  ytick={-0.3,-0.2,-0.1,0,0.1,0.2},
  ymajorgrids, grid style={black!12},
  axis line style={black!40}, tick style={black!40}, x axis line style={draw=none},
  ytick pos=left, xtick style={draw=none},
  extra y ticks={0}, extra y tick labels={}, extra y tick style={grid=major, major grid style={black!70, line width=0.5pt}},
  enlarge x limits=0.09,
  legend style={font=\scriptsize, at={(0.5,1.02)}, anchor=south, legend columns=2, draw=none, /tikz/every even column/.append style={column sep=8pt}},
  legend image code/.code={\draw[#1, draw=none] (0cm,-0.1cm) rectangle (0.25cm,0.1cm);},
  error bars/y dir=both, error bars/y explicit,
  error bars/error bar style={black!65, line width=0.5pt},
]
\addplot[fill=hentblue, draw=none] coordinates {
  (d17,0.086) +- (0,0.022) (d24,0.052) +- (0,0.011) (heb,0.019) +- (0,0.006)
  (g12,-0.018) +- (0,0.014) (q27,0.027) +- (0,0.012) (g31,-0.075) +- (0,0.006)};
\addplot[fill=hentorange, draw=none] coordinates {
  (d17,0.187) +- (0,0.029) (d24,0.075) +- (0,0.008) (heb,0.010) +- (0,0.026)
  (g12,-0.264) +- (0,0.030) (q27,-0.285) +- (0,0.068) (g31,-0.290) +- (0,0.044)};
\legend{\textsc{HEntailment} $\rightarrow$ HebNLI, HebNLI-15k $\rightarrow$ \textsc{HEntailment}}
\end{axis}
\end{tikzpicture}
\caption{Transfer depends on the model. Change in macro-F1 against each decoder's own zero-shot score, in both directions (means over the five sources or test sets). Decoders are ordered by zero-shot score; bars are means of three seeds, whiskers one SD.}
\label{fig:transfer}
\end{figure*}

\paragraph{Stronger models gain less.}
\label{sec:analysis}
Across the six decoders, the mean change on HebNLI after fine-tuning on a \textsc{HEntailment} source falls as zero-shot competence rises (Figure~\ref{fig:transfer}), from $+0.086$ for DictaLM-3.0-1.7B (zero-shot 0.514) to $-0.075$ for gemma-4-31B (0.863); Spearman $\rho = -0.83$, exact one-sided $p = 0.029$, $n = 6$. Leaving out any one decoder keeps the sign ($\rho$ from $-0.70$ to $-1.00$), but without gemma-4-31B or either DictaLM model the correlation is not significant ($p = 0.117$). We therefore report an ordering across these models, not an estimated relationship.

\paragraph{Fine-tuning may shift the neutral boundary.} One possible explanation is that fine-tuning shifts the model's boundary between neutral and contradiction: a model without a reliable inferential prior gains a usable rule, while a model that already separates neutral from entailed loses that separation. We do not test this explanation. The value of the resource depends on the recipient's zero-shot competence.

\paragraph{The loss is in the neutral class.} After HebNLI fine-tuning, gemma-4-12B, gemma-4-31B and Qwen3.6-27B label 59\% to 75\% of neutral \textsc{HEntailment} items as contradictions, against 2\% to 3\% zero-shot (Table~\ref{tab:reverse}). Their neutral F1 falls by 0.42 to 0.62, while their entailment-versus-contradiction accuracy falls by 0.02 to 0.11. These models draw the line between neutral and contradiction differently from \textsc{HEntailment}'s similarity-selected neutrals; we do not test which convention is better.

\paragraph{\textsc{HEntailment} also helps initialize low-resource encoders.} We fine-tune two encoders and two decoders on $N$ HebNLI examples, $N$ from 100 to 15{,}000, either from scratch or starting from the same model fine-tuned on pooled \textsc{HEntailment}. NeoDictaBERT gains 0.21 to 0.31 at 100 to 1{,}000 examples, every seed positive, and 0.03 or less from 5{,}000. XLM-R-large gains 0.25 at 100 examples; its larger apparent gains at 500 and 15{,}000 examples come from unstable from-scratch baselines (seed SD up to 0.33), not from transfer.

\paragraph{It does not help decoders.} Neither decoder shows a consistent gain at any training-set size (DictaLM-3.0-24B $-0.048$ to $+0.011$, gemma-4-31B $-0.015$ to $+0.002$). With two models per architecture we report effect sizes and consistency, not a test (Appendix~\ref{app:lowres}).

\paragraph{The translated benchmarks transfer only partly to each other.} HebNLI-15k raises decoders by 0.074 on LCHAIM against 0.203 on HebNLI itself, and LCHAIM-max lowers their HebNLI score by 0.034, consistent with \citet{talman-chatzikyriakidis-2019-testing}, who find NLI models generalize poorly across benchmarks generally. This caps how much of the repurposed sources' failure can be attributed to repurposing alone (Appendix~\ref{app:lchaim}).

\section{Related Work}
\label{sec:related}

\paragraph{Native NLI beyond English.} A few languages have built NLI natively rather than by translation, by direct annotation \citep{mahendra-etal-2021-indonli, hayashibe-2020-japanese} or by refining an existing corpus \citep{haque2025bnli}. Repurposing sits between the two: cheaper than annotation, native unlike translation.
The concern that translated benchmarks measure something other than native competence is not Hebrew-specific: \citet{kuulmets-fishel-2023-translated} show that a translated Estonian QA test set overestimates performance on native data, \citet{wang-hershcovich-2023-evaluating} that machine translation distorts the semantics of a compositional-generalization benchmark, and \citet{wu2025bitter}, surveying 2{,}000+ multilingual benchmarks, that localized benchmarks align better with human judgments than translated ones (0.68 against 0.47). \citet{thellmann2026quantifying} show that translation errors are associated with measurable accuracy drops on translated benchmarks, and \citet{skorobogat2026roundtrip} that frontier multilingual benchmarks measure mathematical reasoning and factual recall rather than multilingual proficiency. In contrast, \citet{issaka2026translation} find that translation quality predicts downstream multilingual performance, and propose translation as a first-pass screen. Even for translation itself, standard benchmarks saturate and automatic metrics are unreliable, which has led to human-authored hard sets \citep{zouhar2026last}.

\paragraph{Hebrew NLP.} Modern Hebrew is morphologically rich and orthographically ambiguous \citep{tsarfaty2019whats, tsarfaty2020from, keinan2025emotional, keinan2026depression}, and now has the components this method presupposes: encoders \citep{seker-etal-2022-alephbert, shmidman2025neodictabertpushingfrontierbert}, generative models \citep{shmidman-etal-2024-dictalm, shmidman2026dictalm3}, and task datasets beyond those we repurpose \citep{amram-etal-2018-representations, bareket-tsarfaty-2021-neural, paz2023hegel, levy2024parashoot, greenfeld2026beyond, keinan2026here}. Native task datasets without corresponding native NLI resources are a common setting in mid- and low-resource languages \citep{joshi-etal-2020-state}, making them a natural target for our approach.

\paragraph{Auditing constructed NLI.} NLI's label is a relation \emph{between} premise and hypothesis, so a classifier shown the hypothesis alone should be at chance; \citet{poliak-etal-2018-hypothesis} showed across ten corpora that it is not, and \citet{gururangan-etal-2018-annotation} traced this to how hypotheses are elicited. The concern is sharper for constructed data: \citet{proebsting-poliak-2024-hypothesis} report 86--96\% hypothesis-only accuracy on LLM-recreated SNLI against 0.72 on human-written SNLI, which is why we run the check on our own corpus (\S\ref{sec:validation}). This measure characterizes the dataset, not the models trained on it \citep{srikanth-rudinger-2022-partial}, and where that line mitigates artifacts during training \citep{belinkov-etal-2019-dont} we measure them beforehand.

\paragraph{Recasting annotations as inference.} Deriving NLI from other tasks' annotations is established in English, from semantic resources \citep{white-etal-2017-inference}, question answering \citep{wang-etal-2018-glue, demszky2018transforming, trivedi-etal-2019-repurposing} and relation extraction \citep{sainz-etal-2021-label, sainz-etal-2022-textual}. In Hebrew, NLI models fine-tuned on HebNLI have served as denoisers when aligning a knowledge graph to text for relation extraction \citep{keinan2026here}, so Hebrew NLI is already used beyond evaluation. Also, \citet{uppal-etal-2020-two} and \citet{gandhi2024better} extend such recasting to label-scarce task settings and to retrieval over existing datasets, both still within English. WANLI \citep{liu-etal-2022-wanli} is the nearest alternative to repurposing, pairing a generator with human review rather than deriving labels from annotation that already exists. We pursue nativeness in a lower-resource language, and use the result to examine transfer between native and translated NLI in both directions.

\section{Conclusions}
\label{sec:conclusions}

We introduce \textsc{HEntailment}, a native Hebrew NLI benchmark with 160{,}847 training and development pairs and 12{,}613 test pairs, built by repurposing question answering, summarization and paraphrase annotations rather than translating English data.

Transfer between native and translated NLI is neither reliable nor symmetric. Native training helps on translated NLI only for some source tasks and some models, and the strongest models gain least or lose. Translated training helps weaker models on native data but hurts the strongest. Translated and native NLI data are therefore not reliable substitutes for one another.

We release the data, the fine-tuned models and the conversion pipeline. The method needs only native task annotations, a sentence encoder and an LLM fluent in the language, which many mid- and low-resource languages already have. We therefore advocate complementing translated benchmarks with benchmarks derived from native-language tasks.

\section{Limitations}
\label{sec:limitations}

\paragraph{Construction judge as an evaluated model.} gemma-4-31B-it filters the contradictions and is also one of the evaluated decoders. It has the highest zero-shot score and the most negative transfer, and the prior ordering (\S\ref{sec:analysis}) is not significant without it.

\paragraph{Label mix.} Test sets keep the construction's label mix, so contradictions are the minority class (17.6\% of the corpus, 21.2\% of the test pairs, 8\% of QASem's test set). Label-balanced re-scoring changes no conclusion (Appendix~\ref{app:robust}). The neutral-selection ablation uses one encoder.

\paragraph{Domain composition.} Premises inherit the domains of their sources, and any demographic skew those sources carry: Wikipedia and news for HeQ and QASem, news for HeSum and HebSummaries, and the paraphrase corpus's mix. We did not measure demographic composition. The pipeline is source-agnostic, so adding sources can rebalance the benchmark.

\paragraph{Evaluation on Translated Targets.} Ideally, we would evaluate our models against both native and translated benchmarks to isolate the impact of native supervision. However, because this work introduces the first native Hebrew NLI dataset, no independent native test set yet exists.

\section{Ethical Considerations}

\paragraph{Licensing.} Source licenses: HeQ and QASem-Hebrew are CC~BY~4.0, Hebrew-Paraphrase CC~BY-SA~4.0, HeSum and LCHAIM MIT, while HebSummaries and HebNLI carry no explicit license statement. \textsc{HEntailment} is released under CC~BY-SA~4.0 for our annotations and processing; text from a source without a stated license stays under its rights holders' terms. HebNLI is not redistributed.

\paragraph{Annotators.} The validation study was carried out by two students, both native Hebrew speakers, working independently and compensated as part of their paid research positions. No personal data was collected, and the items they annotated are public corpus text.

\paragraph{Inherited Bias.} Repurposed premises from Wikipedia and news media propagate their inherent topical and demographic biases into \textsc{HEntailment}. Benchmark scores should not be interpreted as evidence of robustness across unrepresented groups or domains.

\section*{Acknowledgments}
This research has been funded by the following funding agencies: a grant from the Israeli Science Foundation (ISF), a KAMIN grant by the Israeli Innovation Authority (IIA), and a VATAT grant by the Israeli Planning and Budgeting Committee (PBC), for which we are grateful. We finally wish to thank Google for their generous grant award supporting our lab efforts on AI for Science.

\bibliography{references}

@inproceedings{tsarfaty2019whats,
  title={What's Wrong with Hebrew NLP? And How to Make it Right},
  author={Tsarfaty, Reut and Sadde, Shoval and Klein, Stav and Seker, Amit},
  booktitle={Proceedings of the 2019 Conference on Empirical Methods in Natural Language Processing and the 9th International Joint Conference on Natural Language Processing (EMNLP-IJCNLP): System Demonstrations},
  pages={259--264},
  year={2019}
}

@article{bareket-tsarfaty-2021-neural,
    title={Neural modeling for named entities and morphology (NEMO2)},
    author={Bareket, Dan and Tsarfaty, Reut},
    journal={Transactions of the Association for Computational Linguistics},
    volume={9},
    pages={909--928},
    year={2021},
    publisher={MIT Press}
}

@article{keinan2025emotional,
  title={Emotional analysis in a morphologically rich language: Enhancing machine learning with psychological feature lexicons},
  author={Keinan, Ron and Margalit, Efraim and Bouhnik, Dan},
  journal={Electronics},
  volume={14},
  number={15},
  pages={3067},
  year={2025},
  publisher={MDPI}
}

@article{shmidman-etal-2024-dictalm,
    title={Adapting llms to hebrew: Unveiling dictalm 2.0 with enhanced vocabulary and instruction capabilities},
    author={Shmidman, Shaltiel and Shmidman, Avi and Cohen, Amir DN and Koppel, Moshe},
    journal={arXiv preprint arXiv:2407.07080},
    year={2024}
}

@inproceedings{conneau2018xnli,
  title={XNLI: Evaluating Cross-lingual Sentence Representations},
  author={Conneau, Alexis and Rinott, Ruty and Lample, Guillaume and Williams, Adina and Bowman, Samuel R and Schwenk, Holger and Stoyanov, Veselin},
  booktitle={Proceedings of the 2018 Conference on Empirical Methods in Natural Language Processing},
  pages={2475--2485},
  year={2018}
}

@inproceedings{cohen-etal-2025-hebrewqa,
    title={HeQ: a large and diverse Hebrew reading comprehension benchmark},
    author={Cohen, Amir and Merhav-Fine, Hilla and Goldberg, Yoav and Tsarfaty, Reut},
    booktitle={Findings of the Association for Computational Linguistics: EMNLP 2023},
    pages={13693--13705},
    year={2023}
}

@inproceedings{koppel2011translationese,
  title={Translationese and its dialects},
  author={Koppel, Moshe and Ordan, Noam},
  booktitle={Proceedings of the 49th Annual Meeting of the Association for Computational Linguistics: Human Language Technologies},
  pages={1318--1326},
  year={2011}
}

@inproceedings{levy2024parashoot,
  title     = {{P}ara{S}hoot: A {H}ebrew Question Answering Dataset},
  author    = {Keren, Omri and Levy, Omer},
  booktitle = {Proceedings of the 3rd Workshop on Machine Reading for Question Answering},
  month     = nov,
  year      = {2021},
  address   = {Punta Cana, Dominican Republic},
  publisher = {Association for Computational Linguistics},
  pages     = {106--112},
  doi       = {10.18653/v1/2021.mrqa-1.11},
}

@inproceedings{toral2019post,
  title     = {Post-editese: an Exacerbated Translationese},
  author    = {Toral, Antonio},
  booktitle = {Proceedings of Machine Translation Summit XVII: Research Track},
  pages     = {273--281},
  year      = {2019},
  address   = {Dublin, Ireland},
}

@inproceedings{bowman-etal-2015-large,
    title = "A large annotated corpus for learning natural language inference",
    author = "Bowman, Samuel R.  and
      Angeli, Gabor  and
      Potts, Christopher  and
      Manning, Christopher D.",
    editor = "M{\`a}rquez, Llu{\'i}s  and
      Callison-Burch, Chris  and
      Su, Jian",
    booktitle = "Proceedings of the 2015 Conference on Empirical Methods in Natural Language Processing",
    month = sep,
    year = "2015",
    address = "Lisbon, Portugal",
    publisher = "Association for Computational Linguistics",
    url = "https://aclanthology.org/D15-1075/",
    doi = "10.18653/v1/D15-1075",
    pages = "632--642"
}

@inproceedings{williams-etal-2018-broad,
    title = "A Broad-Coverage Challenge Corpus for Sentence Understanding through Inference",
    author = "Williams, Adina  and
      Nangia, Nikita  and
      Bowman, Samuel R.",
    editor = "Walker, Marilyn  and
      Ji, Heng  and
      Stent, Amanda",
    booktitle = "Proceedings of the 2018 Conference of the North {A}merican Chapter of the Association for Computational Linguistics: Human Language Technologies, Volume 1 (Long Papers)",
    month = jun,
    year = "2018",
    address = "New Orleans, Louisiana",
    publisher = "Association for Computational Linguistics",
    url = "https://aclanthology.org/N18-1101/",
    doi = "10.18653/v1/N18-1101",
    pages = "1112--1122"
}

@inproceedings{artetxe-etal-2020-translation,
    title = "Translation Artifacts in Cross-lingual Transfer Learning",
    author = "Artetxe, Mikel  and
      Labaka, Gorka  and
      Agirre, Eneko",
    editor = "Webber, Bonnie  and
      Cohn, Trevor  and
      He, Yulan  and
      Liu, Yang",
    booktitle = "Proceedings of the 2020 Conference on Empirical Methods in Natural Language Processing (EMNLP)",
    month = nov,
    year = "2020",
    address = "Online",
    publisher = "Association for Computational Linguistics",
    url = "https://aclanthology.org/2020.emnlp-main.618/",
    doi = "10.18653/v1/2020.emnlp-main.618",
    pages = "7674--7684"
}

@article{demszky2018transforming,
  title={Transforming question answering datasets into natural language inference datasets},
  author={Demszky, Dorottya and Guu, Kelvin and Liang, Percy},
  journal={arXiv preprint arXiv:1809.02922},
  year={2018}
}

@inproceedings{wang-etal-2018-glue,
    title = "{GLUE}: A Multi-Task Benchmark and Analysis Platform for Natural Language Understanding",
    author = "Wang, Alex  and
      Singh, Amanpreet  and
      Michael, Julian  and
      Hill, Felix  and
      Levy, Omer  and
      Bowman, Samuel R.",
    editor = "Linzen, Tal  and
      Chrupa{\l}a, Grzegorz  and
      Alishahi, Afra",
    booktitle = "Proceedings of the 2018 {EMNLP} Workshop {B}lackbox{NLP}: Analyzing and Interpreting Neural Networks for {NLP}",
    month = nov,
    year = "2018",
    address = "Brussels, Belgium",
    publisher = "Association for Computational Linguistics",
    url = "https://aclanthology.org/W18-5446/",
    doi = "10.18653/v1/W18-5446",
    pages = "353--355"
}

@inproceedings{dagan2005pascal,
  title={The pascal recognising textual entailment challenge},
  author={Dagan, Ido and Glickman, Oren and Magnini, Bernardo},
  booktitle={Machine learning challenges workshop},
  pages={177--190},
  year={2005},
  organization={Springer}
}

@misc{hebarabnlp2024hebsummaries,
  author       = {{HebArabNlpProject}},
  title        = {HebSummaries},
  year         = {2024},
  publisher    = {Hugging Face},
  howpublished = {\url{https://huggingface.co/datasets/HebArabNlpProject/HebSummaries}}
}

@misc{hebarabnlp2024paraphrase,
  author       = {{HebArabNlpProject}},
  title        = {Hebrew-Paraphrase-Dataset},
  year         = {2025},
  publisher    = {Hugging Face},
  howpublished = {\url{https://huggingface.co/datasets/HebArabNlpProject/Hebrew-Paraphrase-Dataset}}
}

@misc{nnlp2024hebnli,
  author       = {{Israel National NLP Program}},
  title        = {HebNLI},
  year         = {2024},
  publisher    = {Hugging Face},
  howpublished = {\url{https://huggingface.co/datasets/HebArabNlpProject/HebNLI}},
  note         = {Machine translated from the MultiNLI corpus}
}

@inproceedings{madaan-etal-2025-lost,
    title = "Lost in Inference: Rediscovering the Role of Natural Language Inference for Large Language Models",
    author = "Madaan, Lovish  and
      Esiobu, David  and
      Stenetorp, Pontus  and
      Plank, Barbara  and
      Hupkes, Dieuwke",
    editor = "Chiruzzo, Luis  and
      Ritter, Alan  and
      Wang, Lu",
    booktitle = "Proceedings of the 2025 Conference of the Nations of the Americas Chapter of the Association for Computational Linguistics: Human Language Technologies (Volume 1: Long Papers)",
    month = apr,
    year = "2025",
    address = "Albuquerque, New Mexico",
    publisher = "Association for Computational Linguistics",
    url = "https://aclanthology.org/2025.naacl-long.466/",
    doi = "10.18653/v1/2025.naacl-long.466",
    pages = "9229--9242",
    ISBN = "979-8-89176-189-6"
}

@inproceedings{malul-etal-2025-lchaim,
    title = "{LCHAIM} - Investigating Long Context Reasoning in {H}ebrew",
    author = "Malul, Ehud  and
      Perets, Oriel  and
      Mor, Ziv  and
      Kassel, Yigal  and
      Sulem, Elior",
    editor = "Che, Wanxiang  and
      Nabende, Joyce  and
      Shutova, Ekaterina  and
      Pilehvar, Mohammad Taher",
    booktitle = "Findings of the Association for Computational Linguistics: ACL 2025",
    month = jul,
    year = "2025",
    address = "Vienna, Austria",
    publisher = "Association for Computational Linguistics",
    url = "https://aclanthology.org/2025.findings-acl.413/",
    doi = "10.18653/v1/2025.findings-acl.413",
    pages = "7928--7939",
    ISBN = "979-8-89176-256-5"
}

@misc{shmidman2025neodictabertpushingfrontierbert,
      title={NeoDictaBERT: Pushing the Frontier of BERT models for Hebrew}, 
      author={Shaltiel Shmidman and Avi Shmidman and Moshe Koppel},
      year={2025},
      eprint={2510.20386},
      archivePrefix={arXiv},
      primaryClass={cs.CL},
      url={https://arxiv.org/abs/2510.20386}, 
}

@inproceedings{davidov-etal-2026-effective,
    title = "Effective {QA}-Driven Annotation of Predicate{--}Argument Relations Across Languages",
    author = "Davidov, Jonathan  and
      Slobodkin, Aviv  and
      Klein, Shmuel Tomi  and
      Tsarfaty, Reut  and
      Dagan, Ido  and
      Klein, Ayal",
    editor = "Demberg, Vera  and
      Inui, Kentaro  and
      Marquez, Llu{\'i}s",
    booktitle = "Proceedings of the 19th Conference of the {E}uropean Chapter of the {A}ssociation for {C}omputational {L}inguistics (Volume 1: Long Papers)",
    month = mar,
    year = "2026",
    address = "Rabat, Morocco",
    publisher = "Association for Computational Linguistics",
    url = "https://aclanthology.org/2026.eacl-long.112/",
    doi = "10.18653/v1/2026.eacl-long.112",
    pages = "2484--2502",
    ISBN = "979-8-89176-380-7"
}

@article{tsarfaty2020from,
  title={From SPMRL to NMRL: What did we Learn (and Unlearn) in a Decade of Parsing Morphologically-Rich Languages (MRLs)?},
  author={Tsarfaty, Reut and Nivre, Joakim and Andersson, Daniel},
  journal={arXiv preprint arXiv:2005.01330},
  year={2020}
}

@article{sanh2022multitask,
  title={Multitask Prompted Training Enables Zero-Shot Task Generalization},
  author    = {Victor Sanh and Albert Webson and Colin Raffel and Stephen H. Bach and Lintang Sutawika and Zaid Alyafeai and Antoine Chaffin and Arnaud Stiegler and Teven Le Scao and Arun Raja and Manan Dey and M Saiful Bari and Canwen Xu and Urmish Thakker and Shanya Sharma Sharma and Eliza Szczechla and Taewoon Kim and Gunjan Chhablani and Nihal Nayak and Debajyoti Datta and Jonathan Chang and Mike Tian-Jian Jiang and Han Wang and Matteo Manica and Sheng Shen and Zheng Xin Yong and Harshit Pandey and Rachel Bawden and Thomas Wang and Trishala Neeraj and Jos Rozen and Abheesht Sharma and Andrea Santilli and Thibault Fevry and Jason Alan Fries and Ryan Teehan and Tali Bers and Stella Biderman and Leo Gao and Thomas Wolf and Alexander M. Rush},
  journal={arXiv preprint arXiv:2110.08207},
  year={2022}
}

@inproceedings{paz-argaman-etal-2024-hesum,
    title = "{H}e{S}um: a Novel Dataset for Abstractive Text Summarization in {H}ebrew",
    author = "Paz-Argaman, Tzuf  and
      Mondshine, Itai  and
      Achi Mordechai, Asaf  and
      Tsarfaty, Reut",
    editor = "Ku, Lun-Wei  and
      Martins, Andre  and
      Srikumar, Vivek",
    booktitle = "Findings of the Association for Computational Linguistics: ACL 2024",
    month = aug,
    year = "2024",
    address = "Bangkok, Thailand",
    publisher = "Association for Computational Linguistics",
    url = "https://aclanthology.org/2024.findings-acl.381/",
    doi = "10.18653/v1/2024.findings-acl.381",
    pages = "6378--6388",
}

@inproceedings{seker-etal-2022-alephbert,
    title={AlephBERT: Language model pre-training and evaluation from sub-word to sentence level},
    author={Seker, Amit and Bandel, Elron and Bareket, Dan and Brusilovsky, Idan and Greenfeld, Refael and Tsarfaty, Reut},
    booktitle={Proceedings of the 60th Annual Meeting of the Association for Computational Linguistics (Volume 1: Long Papers)},
    pages={46--56},
    year={2022}
}

@article{greenfeld2026beyond,
  title={Beyond Word Boundaries: A Hebrew Coreference Benchmark and an Evaluation Protocol for Morphologically Complex Text},
  author={Greenfeld, Refael Shaked and Tsarfaty, Reut},
  journal={arXiv preprint arXiv:2604.17108},
  year={2026}
}

@inproceedings{kalouli-etal-2019-explaining,
    title = "Explaining Simple Natural Language Inference",
    author = "Kalouli, Aikaterini-Lida  and
      Buis, Annebeth  and
      Real, Livy  and
      Palmer, Martha  and
      de Paiva, Valeria",
    editor = "Friedrich, Annemarie  and
      Zeyrek, Deniz  and
      Hoek, Jet",
    booktitle = "Proceedings of the 13th Linguistic Annotation Workshop",
    month = aug,
    year = "2019",
    address = "Florence, Italy",
    publisher = "Association for Computational Linguistics",
    url = "https://aclanthology.org/W19-4016/",
    doi = "10.18653/v1/W19-4016",
    pages = "132--143"
}

@inproceedings{vanmassenhove-etal-2021-machine,
    title = "Machine Translationese: Effects of Algorithmic Bias on Linguistic Complexity in Machine Translation",
    author = "Vanmassenhove, Eva  and
      Shterionov, Dimitar  and
      Gwilliam, Matthew",
    editor = "Merlo, Paola  and
      Tiedemann, Jorg  and
      Tsarfaty, Reut",
    booktitle = "Proceedings of the 16th Conference of the European Chapter of the Association for Computational Linguistics: Main Volume",
    month = apr,
    year = "2021",
    address = "Online",
    publisher = "Association for Computational Linguistics",
    url = "https://aclanthology.org/2021.eacl-main.188/",
    doi = "10.18653/v1/2021.eacl-main.188",
    pages = "2203--2213"
}

@article{brown2020language,
    title     = {Language Models are Few-Shot Learners},
    author    = {Tom B. Brown and Benjamin Mann and Nick Ryder and Melanie Subbiah and Jared Kaplan and Prafulla Dhariwal and Arvind Neelakantan and Pranav Shyam and Girish Sastry and Amanda Askell and Sandhini Agarwal and Ariel Herbert-Voss and Gretchen Krueger and Tom Henighan and Rewon Child and Aditya Ramesh and Daniel M. Ziegler and Jeffrey Wu and Clemens Winter and Christopher Hesse and Mark Chen and Eric Sigler and Mateusz Litwin and Scott Gray and Benjamin Chess and Jack Clark and Christopher Berner and Sam McCandlish and Alec Radford and Ilya Sutskever and Dario Amodei},
    journal   = {Advances in Neural Information Processing Systems},
    volume    = {33},
    pages     = {1877--1901},
    year      = {2020},
}

@inproceedings{joshi-etal-2020-state,
    title     = {The State and Fate of Linguistic Diversity and Inclusion in the {NLP} World},
    author    = {Joshi, Pratik and Santy, Sebastin and Budhiraja, Amar and Bali, Kalika and Choudhury, Monojit},
    booktitle = {Proceedings of the 58th Annual Meeting of the Association for Computational Linguistics},
    year      = {2020},
    pages     = {6282--6293},
}

@inproceedings{paz2023hegel,
  title={HeGeL: A novel dataset for geo-location from Hebrew text},
  author={Paz-Argaman, Tzuf and Bauman, Tal and Mondshine, Itai and Omer, Itzhak and Dalyot, Sagi and Tsarfaty, Reut},
  booktitle={Findings of the Association for Computational Linguistics: ACL 2023},
  pages={7311--7321},
  year={2023}
}

@article{hu-etal-2022-lora,
    title={Lora: Low-rank adaptation of large language models.},
    author    = {Edward J. Hu and Yelong Shen and Phillip Wallis and Zeyuan Allen-Zhu and Yuanzhi Li and Shean Wang and Lu Wang and Weizhu Chen},
    journal   = {Proceedings of the International Conference on Learning Representations (ICLR)},
    year      = {2022},
}

@inproceedings{sainz-etal-2021-label,
    title     = {Label Verbalization and Entailment for Effective Zero and Few-Shot Relation Extraction},
    author    = {Sainz, Oscar and Lopez de Lacalle, Oier and Labaka, Gorka and Barrena, Ander and Agirre, Eneko},
    booktitle = {Proceedings of the 2021 Conference on Empirical Methods in Natural Language Processing},
    year      = {2021},
    pages     = {1199--1212},
}

@inproceedings{gururangan-etal-2018-annotation,
    title = "Annotation Artifacts in Natural Language Inference Data",
    author = "Gururangan, Suchin and Swayamdipta, Swabha and Levy, Omer and
      Schwartz, Roy and Bowman, Samuel and Smith, Noah A.",
    booktitle = "Proceedings of the 2018 Conference of the North American Chapter of the
      Association for Computational Linguistics: Human Language Technologies, Volume 2 (Short Papers)",
    year = "2018",
    address = "New Orleans, Louisiana",
    publisher = "Association for Computational Linguistics",
    pages = "107--112",
}

@inproceedings{poliak-etal-2018-hypothesis,
    title = "Hypothesis Only Baselines in Natural Language Inference",
    author = "Poliak, Adam and Naradowsky, Jason and Haldar, Aparajita and
      Rudinger, Rachel and Van Durme, Benjamin",
    booktitle = "Proceedings of the Seventh Joint Conference on Lexical and Computational Semantics",
    year = "2018",
    address = "New Orleans, Louisiana",
    publisher = "Association for Computational Linguistics",
    pages = "180--191",
}

@inproceedings{white-etal-2017-inference,
    title = "Inference is Everything: Recasting Semantic Resources into a Unified Evaluation Framework",
    author = "White, Aaron Steven  and
      Rastogi, Pushpendre  and
      Duh, Kevin  and
      Van Durme, Benjamin",
    editor = "Kondrak, Greg  and
      Watanabe, Taro",
    booktitle = "Proceedings of the Eighth International Joint Conference on Natural Language Processing (Volume 1: Long Papers)",
    month = nov,
    year = "2017",
    address = "Taipei, Taiwan",
    publisher = "Asian Federation of Natural Language Processing",
    url = "https://aclanthology.org/I17-1100/",
    pages = "996--1005"
}

@inproceedings{trivedi-etal-2019-repurposing,
    title = "Repurposing Entailment for Multi-Hop Question Answering Tasks",
    author = "Trivedi, Harsh  and
      Kwon, Heeyoung  and
      Khot, Tushar  and
      Sabharwal, Ashish  and
      Balasubramanian, Niranjan",
    editor = "Burstein, Jill  and
      Doran, Christy  and
      Solorio, Thamar",
    booktitle = "Proceedings of the 2019 Conference of the North {A}merican Chapter of the Association for Computational Linguistics: Human Language Technologies, Volume 1 (Long and Short Papers)",
    month = jun,
    year = "2019",
    address = "Minneapolis, Minnesota",
    publisher = "Association for Computational Linguistics",
    url = "https://aclanthology.org/N19-1302/",
    doi = "10.18653/v1/N19-1302",
    pages = "2948--2958"
}

@inproceedings{uppal-etal-2020-two,
    title = "Two-Step Classification using Recasted Data for Low Resource Settings",
    author = "Uppal, Shagun  and
      Gupta, Vivek  and
      Swaminathan, Avinash  and
      Zhang, Haimin  and
      Mahata, Debanjan  and
      Gosangi, Rakesh  and
      Shah, Rajiv Ratn  and
      Stent, Amanda",
    editor = "Wong, Kam-Fai  and
      Knight, Kevin  and
      Wu, Hua",
    booktitle = "Proceedings of the 1st Conference of the Asia-Pacific Chapter of the Association for Computational Linguistics and the 10th International Joint Conference on Natural Language Processing",
    month = dec,
    year = "2020",
    address = "Suzhou, China",
    publisher = "Association for Computational Linguistics",
    url = "https://aclanthology.org/2020.aacl-main.71/",
    doi = "10.18653/v1/2020.aacl-main.71",
    pages = "706--719"
}

@inproceedings{kuulmets-fishel-2023-translated,
    title = "Translated Benchmarks Can Be Misleading: the Case of {E}stonian Question Answering",
    author = "Kuulmets, Hele-Andra  and
      Fishel, Mark",
    editor = {Alum{\"a}e, Tanel  and
      Fishel, Mark},
    booktitle = "Proceedings of the 24th Nordic Conference on Computational Linguistics (NoDaLiDa)",
    month = may,
    year = "2023",
    address = "T{\'o}rshavn, Faroe Islands",
    publisher = "University of Tartu Library",
    url = "https://aclanthology.org/2023.nodalida-1.71/",
    pages = "710--716"
}

@inproceedings{wang-hershcovich-2023-evaluating,
    title = "On Evaluating Multilingual Compositional Generalization with Translated Datasets",
    author = "Wang, Zi  and
      Hershcovich, Daniel",
    editor = "Rogers, Anna  and
      Boyd-Graber, Jordan  and
      Okazaki, Naoaki",
    booktitle = "Proceedings of the 61st Annual Meeting of the Association for Computational Linguistics (Volume 1: Long Papers)",
    month = jul,
    year = "2023",
    address = "Toronto, Canada",
    publisher = "Association for Computational Linguistics",
    url = "https://aclanthology.org/2023.acl-long.93/",
    doi = "10.18653/v1/2023.acl-long.93",
    pages = "1669--1687"
}

@inproceedings{srikanth-rudinger-2022-partial,
    title = "Partial-input baselines show that {NLI} models can ignore context, but they don{'}t.",
    author = "Srikanth, Neha  and
      Rudinger, Rachel",
    editor = "Carpuat, Marine  and
      de Marneffe, Marie-Catherine  and
      Meza Ruiz, Ivan Vladimir",
    booktitle = "Proceedings of the 2022 Conference of the North American Chapter of the Association for Computational Linguistics: Human Language Technologies",
    month = jul,
    year = "2022",
    address = "Seattle, United States",
    publisher = "Association for Computational Linguistics",
    url = "https://aclanthology.org/2022.naacl-main.350/",
    doi = "10.18653/v1/2022.naacl-main.350",
    pages = "4753--4763"
}

@inproceedings{liu-etal-2022-wanli,
    title = "{WANLI}: Worker and {AI} Collaboration for Natural Language Inference Dataset Creation",
    author = "Liu, Alisa  and
      Swayamdipta, Swabha  and
      Smith, Noah A.  and
      Choi, Yejin",
    editor = "Goldberg, Yoav  and
      Kozareva, Zornitsa  and
      Zhang, Yue",
    booktitle = "Findings of the Association for Computational Linguistics: EMNLP 2022",
    month = dec,
    year = "2022",
    address = "Abu Dhabi, United Arab Emirates",
    publisher = "Association for Computational Linguistics",
    url = "https://aclanthology.org/2022.findings-emnlp.508/",
    doi = "10.18653/v1/2022.findings-emnlp.508",
    pages = "6826--6847"
}

@inproceedings{mahendra-etal-2021-indonli,
    title = "{I}ndo{NLI}: A Natural Language Inference Dataset for {I}ndonesian",
    author = "Mahendra, Rahmad  and
      Aji, Alham Fikri  and
      Louvan, Samuel  and
      Rahman, Fahrurrozi  and
      Vania, Clara",
    editor = "Moens, Marie-Francine  and
      Huang, Xuanjing  and
      Specia, Lucia  and
      Yih, Scott Wen-tau",
    booktitle = "Proceedings of the 2021 Conference on Empirical Methods in Natural Language Processing",
    month = nov,
    year = "2021",
    address = "Online and Punta Cana, Dominican Republic",
    publisher = "Association for Computational Linguistics",
    url = "https://aclanthology.org/2021.emnlp-main.821/",
    doi = "10.18653/v1/2021.emnlp-main.821",
    pages = "10511--10527"
}

@inproceedings{hayashibe-2020-japanese,
    title = "{J}apanese Realistic Textual Entailment Corpus",
    author = "Hayashibe, Yuta",
    editor = "Calzolari, Nicoletta  and
      B{\'e}chet, Fr{\'e}d{\'e}ric  and
      Blache, Philippe  and
      Choukri, Khalid  and
      Cieri, Christopher  and
      Declerck, Thierry  and
      Goggi, Sara  and
      Isahara, Hitoshi  and
      Maegaard, Bente  and
      Mariani, Joseph  and
      Mazo, H{\'e}l{\`e}ne  and
      Moreno, Asuncion  and
      Odijk, Jan  and
      Piperidis, Stelios",
    booktitle = "Proceedings of the Twelfth Language Resources and Evaluation Conference",
    month = may,
    year = "2020",
    address = "Marseille, France",
    publisher = "European Language Resources Association",
    url = "https://aclanthology.org/2020.lrec-1.843/",
    pages = "6827--6834",
    language = "eng",
    ISBN = "979-10-95546-34-4"
}

@inproceedings{talman-chatzikyriakidis-2019-testing,
    title = "Testing the Generalization Power of Neural Network Models across {NLI} Benchmarks",
    author = "Talman, Aarne  and
      Chatzikyriakidis, Stergios",
    editor = "Linzen, Tal  and
      Chrupa{\l}a, Grzegorz  and
      Belinkov, Yonatan  and
      Hupkes, Dieuwke",
    booktitle = "Proceedings of the 2019 ACL Workshop BlackboxNLP: Analyzing and Interpreting Neural Networks for NLP",
    month = aug,
    year = "2019",
    address = "Florence, Italy",
    publisher = "Association for Computational Linguistics",
    url = "https://aclanthology.org/W19-4810/",
    doi = "10.18653/v1/W19-4810",
    pages = "85--94"
}

@inproceedings{sainz-etal-2022-textual,
    title = "Textual Entailment for Event Argument Extraction: Zero- and Few-Shot with Multi-Source Learning",
    author = "Sainz, Oscar  and
      Gonzalez-Dios, Itziar  and
      Lopez de Lacalle, Oier  and
      Min, Bonan  and
      Agirre, Eneko",
    editor = "Carpuat, Marine  and
      de Marneffe, Marie-Catherine  and
      Meza Ruiz, Ivan Vladimir",
    booktitle = "Findings of the Association for Computational Linguistics: NAACL 2022",
    month = jul,
    year = "2022",
    address = "Seattle, United States",
    publisher = "Association for Computational Linguistics",
    url = "https://aclanthology.org/2022.findings-naacl.187/",
    doi = "10.18653/v1/2022.findings-naacl.187",
    pages = "2439--2455"
}

@article{gandhi2024better,
  title     = {Better Synthetic Data by Retrieving and Transforming Existing Datasets},
  author    = {Saumya Gandhi and Ritu Gala and Vijay Viswanathan and Tongshuang Wu and Graham Neubig},
  year      = {2024},
  journal   = {arXiv preprint arXiv:2404.14361},
  eprint    = {2404.14361},
  archivePrefix = {arXiv},
}

@article{wu2025bitter,
  title     = {The Bitter Lesson Learned from 2,000+ Multilingual Benchmarks},
  author    = {Minghao Wu and Weixuan Wang and Sinuo Liu and Huifeng Yin and Xintong Wang and Yu Zhao and Chenyang Lyu and Longyue Wang and Weihua Luo and Kaifu Zhang},
  year      = {2025},
  journal   = {arXiv preprint arXiv:2504.15521},
  eprint    = {2504.15521},
  archivePrefix = {arXiv},
}

@article{skorobogat2026roundtrip,
  title     = {Round-Trip Translation Reveals What Frontier Multilingual Benchmarks Miss},
  author    = {Ronald Skorobogat and Ameya Prabhu and Matthias Bethge},
  year      = {2026},
  journal   = {arXiv preprint arXiv:2604.12911},
  eprint    = {2604.12911},
  archivePrefix = {arXiv}
}

@article{issaka2026translation,
  title     = {Translation as a Scalable Proxy for Multilingual Evaluation},
  author    = {Sheriff Issaka and Erick Rosas Gonzalez and Lieqi Liu and Evans Kofi Agyei and Lucas Bandarkar and Nanyun Peng and David Ifeoluwa Adelani and Francisco Guzm{\'a}n and Saadia Gabriel},
  year      = {2026},
  journal   = {arXiv preprint arXiv:2601.11778},
  eprint    = {2601.11778},
  archivePrefix = {arXiv},
}

@article{thellmann2026quantifying,
  title     = {Quantifying the Impact of Translation Errors on Multilingual LLM Evaluation},
  author    = {Klaudia-Doris Thellmann and Bernhard Stadler and Michael Färber and Jens Lehmann},
  year      = {2026},
  journal   = {arXiv preprint arXiv:2605.24904},
  eprint    = {2605.24904},
  archivePrefix = {arXiv}
}

@inproceedings{proebsting-poliak-2024-hypothesis,
  title     = {Biases in Large Language Model-Elicited Text: A Case Study in Natural Language Inference},
  author    = {Proebsting, Grace and Poliak, Adam},
  booktitle = {Proceedings of the 31st International Conference on Computational Linguistics},
  year      = {2025},
  address   = {Abu Dhabi, UAE},
  publisher = {Association for Computational Linguistics},
  url       = {https://aclanthology.org/2025.coling-main.389/}
}

@inproceedings{belinkov-etal-2019-dont,
    title = "Don{'}t Take the Premise for Granted: Mitigating Artifacts in Natural Language Inference",
    author = "Belinkov, Yonatan  and
      Poliak, Adam  and
      Shieber, Stuart  and
      Van Durme, Benjamin  and
      Rush, Alexander",
    editor = "Korhonen, Anna  and
      Traum, David  and
      M{\`a}rquez, Llu{\'i}s",
    booktitle = "Proceedings of the 57th Annual Meeting of the Association for Computational Linguistics",
    month = jul,
    year = "2019",
    address = "Florence, Italy",
    publisher = "Association for Computational Linguistics",
    url = "https://aclanthology.org/P19-1084/",
    doi = "10.18653/v1/P19-1084",
    pages = "877--891"
}

@article{haque2025bnli,
  title     = {BNLI: A Linguistically-Refined Bengali Dataset for Natural Language Inference},
  author    = {Farah Binta Haque and Md Yasin and Shishir Saha and Md Shoaib Akhter Rafi},
  year      = {2025},
  journal   = {arXiv preprint arXiv:2511.08813},
  eprint    = {2511.08813},
  archivePrefix = {arXiv}
}

@misc{shmidman2026dictalm3,
  title         = {Dicta-LM 3.0: Advancing The Frontier of Hebrew Sovereign LLMs},
  author        = {Shmidman, Shaltiel and Shmidman, Avi and Cohen, Amir DN and Koppel, Moshe},
  year          = {2026},
  eprint        = {2602.02104},
  archivePrefix = {arXiv},
  primaryClass  = {cs.CL},
  url           = {https://arxiv.org/abs/2602.02104},
}

@inproceedings{amram-etal-2018-representations,
    title = "Representations and Architectures in Neural Sentiment Analysis for Morphologically Rich Languages: A Case Study from {M}odern {H}ebrew",
    author = "Amram, Adam and Ben David, Anat and Tsarfaty, Reut",
    booktitle = "Proceedings of the 27th International Conference on Computational Linguistics",
    month = aug,
    year = "2018",
    address = "Santa Fe, New Mexico, USA",
    publisher = "Association for Computational Linguistics",
    url = "https://aclanthology.org/C18-1190/",
    pages = "2242--2252"
}

@inproceedings{mor-lan-etal-2026-location,
    title = "Location Not Found: Exposing Implicit Local and Global Biases in Multilingual {LLM}s",
    author = "Mor-Lan, Guy  and
      Goldman, Omer  and
      Eyal, Matan  and
      Gilady, Adi Mayrav  and
      Eiger, Sivan  and
      Szpektor, Idan  and
      Hassidim, Avinatan  and
      Matias, Yossi  and
      Tsarfaty, Reut",
    editor = "Liakata, Maria  and
      Moreira, Viviane P.  and
      Zhang, Jiajun  and
      Jurgens, David",
    booktitle = "Proceedings of the 64th Annual Meeting of the {A}ssociation for {C}omputational {L}inguistics (Volume 1: Long Papers)",
    month = jul,
    year = "2026",
    address = "San Diego, California, United States",
    publisher = "Association for Computational Linguistics",
    url = "https://aclanthology.org/2026.acl-long.1344/",
    doi = "10.18653/v1/2026.acl-long.1344",
    pages = "29130--29150",
    ISBN = "979-8-89176-390-6"
}

@article{dagan2013recognizing,
  title={Recognizing textual entailment: Models and applications},
  author={Dagan, Ido and Roth, Dan and Sammons, Mark and Zanzotto, Fabio Massimo},
  journal={Synthesis Lectures on Human Language Technologies},
  volume={6},
  number={4},
  year={2013}
}

@misc{qwen2026qwen36,
  title        = {Qwen3.6-27B},
  author       = {{Qwen Team}},
  year         = {2026},
  howpublished = {\url{https://huggingface.co/Qwen/Qwen3.6-27B}},
}

@misc{gemma2026gemma4,
  title        = {Gemma 4 Model Card},
  author       = {{Gemma Team}},
  year         = {2026},
  howpublished = {\url{https://huggingface.co/google/gemma-4-31B-it}},
}

@misc{hebatron2026,
  title        = {Hebatron},
  author       = {{HebArabNlpProject}},
  year         = {2026},
  howpublished = {\url{https://huggingface.co/HebArabNlpProject/Hebatron}},
}

@inproceedings{keinan2026here,
  title     = {{HeRE}: A Novel Benchmark for {H}ebrew Relation Extraction via {LLM}-Guided Denoising of Knowledge Graph Alignments},
  author    = {Keinan, Ron and Cohen, Amir David Nissan and Tsarfaty, Reut},
  booktitle = {Proceedings of the 15th International Joint Conference on Natural Language Processing and the 5th Conference of the Asia-Pacific Chapter of the Association for Computational Linguistics (Volume 1: Long Papers)},
  pages     = {5311--5329},
  year      = {2026},
}

@misc{zouhar2026last,
  title         = {Last Translation Benchmark},
  author    = {Vil{\'e}m Zouhar and Niyati Bafna and Mukund Choudhary and Maike Z{\"u}fle and Sara Rajaee and Pinzhen Chen and Jannis Vamvas and Sara Papi and Ona de Gibert and Bhavitvya Malik and Eliya Habba and Orfeas Menis Mastromichalakis and Patr{\'i}cia Schmidtov{\'a} and Michelle Wastl and Sheriff Issaka and Leshem Choshen and Stella Biderman and Antonis Anastasopoulos and Jan Niehues and Rico Sennrich and Mrinmaya Sachan and Ond{\v r}ej Bojar and Kenton Murray and J{\"o}rg Tiedemann and Alham Fikri Aji and Philipp Koehn and Christof Monz and Alexandra Birch and Sowmya Vajjala and Chalamalasetti Kranti and Cristina Espa{\~n}a-Bonet and Nobin Sarwar and David Kacz{\'e}r and Sourajit Saha and Jonathan Tonglet and Shunta Asano and Malik Marmonier and Daban Q. Jaff and Vaisakhi Mishra and Hend Al-Khalifa and Gabriele Sarti and Nils Rehlinger and Juan Daniel Cuervo Villa and Dominik Mach{\'a}{\v c}ek and Saugata Purkayastha and Jagannathan Ramanujam and Shubhashis Roy Dipta and Aviral Nigam and Shuaib Shuaib Yusuf and Heejin Do and Jonathan Yahav and Johannes-Rudolf David and Maria Carmen Staiano and Zuzana Nadova and Fred Philippy and Ron Keinan and Maria Lymperaiou and Silvia Casola and Fabian Retkowski and Andr{\'e}s Jerez and Hanna Yukhy- menko and Sangwon Ryu and Avantica Vempati and Sukannya Purkayastha and Adrian Cosma and Erivan Inan and Vitalii Babenko and Wafa Aissa and Valentin Scourneau and Fatima Haouari and Venkata Prasanth Kumar Gummadi and Mehdi Jafarzadeh and Manon Reusens and Kaiser Sun and Lukas Edman and Shaomu Tan and Giuseppe Gallipoli and Pawan Sasanka Ammanamanchi and Manar Ali and Mohammad Sadegh Gholizadeh and Dipankar Srirag and Marek {\v S}uppa and Javier Garc{\'i}a Gilabert and Ruta Binkyte and Ana-Maria Bucur and Sabry E. Farrag and Youssef Saber and Yihong Liu and Theresia Veronika Rampisela and Christian Hoang and Nicoleta Cojocaru and Jan Koco{\'n} and Jean Maillard and Xiaochuang Yuan and Sina Ahmadi and Daryna Dementieva and Philipp Mondorf and Kaustubh Dhole and Valmik Nahata and Roman Wixinger and Amir Hossein Yari and Shenbin Qian and Manuel Tuor and Fida Mohammad Thoker and Sergey Troshin and Lance Calvin Lim Gamboa and Amir Arsalan Rezapour and K{\"a}triin Kukk and Koel Dutta Chowdhury and Shaswati Saha and Seth Aycock and Bo Chen and Linh Vu and Vatsal Venkatkrishna and Shayan Bali and Arafat Ahsan and Luan Thanh Nguyen and Hassan Soliman and Ngoc Quynh Tram Do and Azmine Toushik Wasi and R. Damanhuri and Marius Huber and Kazuki Egashira and Jimson Paulo Layacan and David Africa and Vladislav Poritski and Mike Zhang and Deep Shah and Abdulaziz Nura Kani and Luis Frentzen Salim and Paul Gavrikov and Bello Umar Bello and Ayush Sunil Munot and Anumit Garg and Yolanda Xavier and Qiaoyuan Zheng and Kawsar Ahmed and Debanshu Das and Zimu Wang and Gengyu Rao and Kamile Dementaviciute and Farhan Farsi and L D M S Sai Teja and Dawei Zhu and Yi Fan and Wei Liu and Marco Gaido and Elias Herranen and Sankalan Pal Chowdhury and Karen Sanchez and Guy Kaplan and Farzad Shami and Ashok Urlana and Amir Hossein Kargaran and Sofie Goethals and Priyaranjan Pattnayak and Oksana Volchek and Marii Ojastu and Hongbin Na and Emilian Radoi and Chenyi Zhao and Carlos Hinojosa and Andrei Niculae and Andrea Gregor de Varda and Zaid Alyafeai and Tomasz Limisiewicz and Reem Alzahrani and Pouya Sadeghi and Nehal Kathrotia and Mateusz Lango and Enzo Doyen and Alex Fl{\"u}ckiger and Ulysses Sekai Tully Carr and Samuel Simko and Ritwik Tiwari and Rishit Dagli and Isaac R Caswell and Bowen Yi and Aicha Chorana and Selja Ker{\"a}nen and Sadiksha Chitrakar and Muhammad Ravi Shulthan Habibi and Joy Olusanya and Bishal Shrestha and Zhengxiang Wang and Vivek Harsha Lakkamaneni and Sophia Conrad and Panayiotis Panayiotou and Nazia Tasnim and Marta Punsola Mun{\'a}rriz and Marko Culjak and Luis Lara and Jenny Chim and Jannatul Nayem and Fidel Rodr{\'i}guez Vel{\'a}squez and Eran Yahav and Blanka K{\"o}v{\'e}r and Beatrice Savoldi and Anmol Goel and Aishik Mandal and Tosin Adewumi and Raoyuan Zhao and Mykola Haltiuk and Antonia Karamolegkou and Yuxing Lu and Thura Aung and Naser Almousa and Tommaso Cerruti and Raia Abu Ahmad and Beni Egressy and Alireza Pakniat and St{\'e}phane J. P. S. Thunus and Rachel Bawden and Lena Libon and Samridhya Biswas and Prakhar Gupta and Nusrat Jahan Lia and Nguyen Tai and Natchapon Jongwiriyanurak and Minh Ngoc Do and Ivan Bara{\'c} and Dzmitry Kuzmin and Badal Nyalang and Antoine Taroni and Andy Catruna and Rushikesh Zawar and Roland Aydin and Pavel Stepachev and Ilai Yaron Levy and Andreas Simons and Rayyan Merchant and Ziyi Yang and Samuel Frontull and Kenneth Enevoldsen and Harris Abdul Majid and Tim Graf and Tatiana Bielakova and Sharifa Djurabaeva and Shaoxiong Ji and Jirui Qi and Ayla Rigouts Terryn and Yurii Paniv and Xiyan Fu and Sunisth Kumar and Shree Harsha Bokkahalli Satish and Papa Abdou Karim Karou Diallo and Mengyu Ye and Maximilian Vieweg and Matija Akrap and Krist{\'y}na Onderkov{\'a} and Joseph Attieh and Ivan Yuri De Leon and Ibrahim Baroud and Esrael Teferi Tensay and Elisabeth Fittschen and David Duki{\'c} and Beno{\^i}t Sagot and Jingwei Ni and Yu Fan and Juri Opitz},
  year          = {2026},
  eprint        = {2609.04173},
  archivePrefix = {arXiv},
  primaryClass  = {cs.CL},
}

@misc{shizgal2026dqi,
  title         = {From {DQI} to {C-DQI}: Automatic Codebook Generation and Evaluation for Measuring Computational Discourse Quality Index},
  author    = {Esther Shizgal and Ron Keinan and Daniel Shtauber and Roee Tzvi Esquira and Ori Aronson and Matan Toledano and Or Rappel-Kroyzer and Liron Lavi and Renana Keydar and Ittai Bar-Siman-Tov and Reut Tsarfaty},
  year          = {2026},
  eprint        = {2610.27937},
  archivePrefix = {arXiv},
  primaryClass  = {cs.CL},
}

@inproceedings{dettmers2022optimizers,
  title     = {8-bit Optimizers via Block-wise Quantization},
  author    = {Dettmers, Tim and Lewis, Mike and Shleifer, Sam and Zettlemoyer, Luke},
  booktitle = {International Conference on Learning Representations},
  year      = {2022},
}

@inproceedings{dettmers2023qlora,
  title     = {{QLoRA}: Efficient Finetuning of Quantized {LLM}s},
  author    = {Dettmers, Tim and Pagnoni, Artidoro and Holtzman, Ari and Zettlemoyer, Luke},
  booktitle = {Advances in Neural Information Processing Systems},
  volume    = {36},
  year      = {2023},
}

@inproceedings{kwon2023efficient,
  title     = {Efficient Memory Management for Large Language Model Serving with {PagedAttention}},
  author    = {Kwon, Woosuk and Li, Zhuohan and Zhuang, Siyuan and Sheng, Ying and Zheng, Lianmin and Yu, Cody Hao and Gonzalez, Joseph E. and Zhang, Hao and Stoica, Ion},
  booktitle = {Proceedings of the 29th Symposium on Operating Systems Principles},
  pages     = {611--626},
  year      = {2023},
}

@misc{peft,
  title        = {{PEFT}: State-of-the-art Parameter-Efficient Fine-Tuning Methods},
  author       = {Mangrulkar, Sourab and Gugger, Sylvain and Debut, Lysandre and Belkada, Younes and Paul, Sayak and Bossan, Benjamin},
  year         = {2022},
  howpublished = {\url{https://github.com/huggingface/peft}},
}

@inproceedings{lebras2020adversarial,
  title     = {Adversarial Filters of Dataset Biases},
  author    = {Le Bras, Ronan and Swayamdipta, Swabha and Bhagavatula, Chandra and Zellers, Rowan and Peters, Matthew and Sabharwal, Ashish and Choi, Yejin},
  booktitle = {Proceedings of the 37th International Conference on Machine Learning},
  series    = {Proceedings of Machine Learning Research},
  volume    = {119},
  pages     = {1078--1088},
  year      = {2020},
}

@article{cohen1960coefficient,
  title   = {A Coefficient of Agreement for Nominal Scales},
  author  = {Cohen, Jacob},
  journal = {Educational and Psychological Measurement},
  volume  = {20},
  number  = {1},
  pages   = {37--46},
  year    = {1960},
}

@article{keinan2026depression,
  title     = {Depression Detection with Machine Learning in a Low-Resource Language: A Case Study in {H}ebrew Digital Health Communities},
  author    = {Keinan, Ron and Margalit, Efraim and Bouhnik, Dan},
  journal   = {Array},
  pages     = {101157},
  year      = {2026},
  publisher = {Elsevier},
}

@inproceedings{wei2022finetuned,
  title     = {Finetuned Language Models are Zero-Shot Learners},
  author    = {Wei, Jason and Bosma, Maarten and Zhao, Vincent Y. and Guu, Kelvin and Yu, Adams Wei and Lester, Brian and Du, Nan and Dai, Andrew M. and Le, Quoc V.},
  booktitle = {International Conference on Learning Representations},
  year      = {2022},
}

\appendix

\section{Construction Details}
\label{app:strategies}

\paragraph{Paraphrase is bidirectional.} Hebrew-Paraphrase pairs are mutually entailing, so both directions (original~$\rightarrow$~paraphrase and paraphrase~$\rightarrow$~original) are emitted where both pass the filters. Neutral partners are drawn within the same length bucket (sentence with sentence, paragraph with paragraph), so the pairing adds no length cue.

\paragraph{Contradiction rules.} Every rule edits the entailed hypothesis. Number edits replace a number with a plausible one of the same format: years shift by 1 to 9, percentages stay between 1 and 100, other numbers keep their digit count. The QA corpora include distractor answers; the pipeline does not use them. We rank candidate edits, putting single-token changes to a detail the premise states first, and send the top candidates to the judge (gemma-4-31B-it, greedy), which answers two questions: is the edited sentence grammatical Hebrew, and does the premise contradict it. The first candidate that passes both is kept. A contradiction that edits a detail absent from the premise (\emph{ungrounded}) is kept only when no grounded candidate passes; ungrounded items are 5.5\% (HebSummaries) to 27.2\% (HeQ) of each source's contradictions, and every contradiction carries the flag. By primary rule, the 30{,}467 contradictions are entity swaps (person 27\%, organization 25\%, place 14\%), number edits (16\%), adjective antonyms (5\%), superlative flips (3\%), quantifier flips (3\%), and decision inversion, state transition, causal reversal and other lexical rules (6\%). Table~\ref{tab:funnel} gives the yield per source.

\paragraph{Excluded rules.} For each rule we train a hypothesis-only probe to separate its contradictions from entailments, with folds grouped by premise. Six rules leave their label in the edited word and are excluded: verb negation (0.98 on HeQ), modal flip (0.96), existential negation (0.95), temporal inversion (0.94), time-expression swap (0.82) and epistemic certainty. The kept rules read 0.52 (person swap) to 0.71 (adjective antonym) against their own entailment twins, chance 0.50. The number rule keeps the number's format.

\begin{table}[h]
\centering\small
\setlength{\tabcolsep}{4pt}
\begin{tabular}{lrrcc}
\toprule
\textbf{Source} & \textbf{Ent.} & \textbf{Con.} & \textbf{Yield} & \textbf{Ungr.} \\
\midrule
HeQ          & 21{,}784 & 9{,}156 & 0.42 & 27.2\% \\
QASem        & 21{,}157 & 3{,}128 & 0.15 & 19.0\% \\
HeSum        &  9{,}997 & 3{,}897 & 0.39 &  7.7\% \\
HebSumm.     &  5{,}024 & 4{,}328 & 0.86 &  5.5\% \\
HebPara.  & 17{,}382 & 9{,}958 & 0.57 & 22.5\% \\
\midrule
\textbf{All} & \textbf{75{,}344} & \textbf{30{,}467} & \textbf{0.40} & \\
\bottomrule
\end{tabular}
\caption{Contradiction yield per entailment hypothesis, and the share of ungrounded contradictions.}
\label{tab:funnel}
\end{table}

\begin{figure}[h]
\centering\small
\begin{tabular}{@{}p{0.95\columnwidth}@{}}
\toprule
\textbf{Premise}: \emph{...the prices rose sharply, so that the inflation rate in \textbf{1952} reached 66\%...} \\[3pt]
\textbf{Entailment}: \emph{The inflation rate in Israel in \textbf{1952} was 66\%.} \\[3pt]
\textbf{Contradiction}: \emph{The inflation rate in Israel in \underline{\textbf{1949}} was 66\%.} \\
\bottomrule
\end{tabular}
\caption{A HeQ contradiction and its entailment twin (translated). The number rule shifts a year by 1 to 9.}
\label{fig:twin}
\end{figure}

\subsection{Generation Provenance}
\label{app:genprov}

\paragraph{LLMs in construction.} \texttt{gemini-3.1-flash-lite} turns a (question, gold answer) pair into a declarative sentence for HeQ and QASem, and de-segments QASem's text; the other three sources use source text directly. gemma-4-31B-it, run locally, filters the contradictions of all five sources. Qwen3.6-27B, also local, is the second judge of \S\ref{sec:validation} and takes no part in construction. Agreement between independent LLM judges has been used as a quality signal for Hebrew parliamentary text \citep{shizgal2026dqi}. All prompts are in the release.

\paragraph{Decoding and caching.} Gemini is called through the OpenRouter API with a single user message and no sampling parameters, so decoding follows the provider's defaults, which we do not control. Failed calls are retried three times and then dropped (62 of 64{,}106 calls, all de-segmentation). Every response is cached by source, split and item, and the caches are released, so the corpus can be rebuilt without new calls. The local judges decode greedily. The Gemini prompt, translated from Hebrew:

\begin{samepage}
\begin{quote}\small\ttfamily
The following is a question and answer in Hebrew. Write one declarative sentence in standard Hebrew that expresses the answer to the question. Return only the sentence, with no explanation.\\
Question: \{question\}\\
Answer: \{answer\}
\end{quote}
\end{samepage}

\paragraph{Contamination.} All five source corpora are public and predate the evaluated models, so they may be in pretraining data. This bears on absolute zero-shot scores more than on transfer deltas, which compare each model against itself; we have not measured it.

\section{Ablation: Neutral Selection}
\label{app:revised}

A neutral hypothesis comes from a different document, but which one is a design choice. Drawing from the most \emph{distant} candidates makes neutral pairs topically disjoint from their premises. \textsc{HEntailment} draws from the most \emph{similar} candidates below the entailment ceiling of 0.7, which gives neutral pairs more overlap with the premise and moves them toward HebNLI's. Table~\ref{tab:neutralabl} trains XLM-R-large on each source with either policy (three seeds). Similar neutrals stay learnable: they score as well or better on four of five sources. The HeSum difference comes from two distant-neutral runs that stayed at chance (macro-F1 0.26 and 0.24, against 0.52 for the third seed). With one encoder we do not claim an effect of the policy on transfer.

\begin{table}[h]
\centering\small
\begin{tabular}{lccc}
\toprule
\textbf{Source} & \textbf{distant} & \textbf{similar} & \textbf{$\Delta$} \\
\midrule
HeQ               & 0.924 & 0.963 & $-0.039$ \\
QASem             & 0.711 & 0.888 & $-0.177$ \\
HeSum             & 0.339 & 0.844 & $-0.505$ \\
HebSumm.      & 0.732 & 0.606 & $+0.125$ \\
HebPara. & 0.897 & 0.912 & $-0.015$ \\
\bottomrule
\end{tabular}
\caption{Own-source macro-F1 of XLM-R-large with distant or similar neutrals (the released corpus uses similar), mean of three seeds.}
\label{tab:neutralabl}
\end{table}

\section{Applicability}

The pipeline presupposes annotated data whose labels encode inferential structure, plus a sentence encoder and an LLM fluent in the target language. That combination is more widely available than native NLI but is not universal, and languages lacking all three, often those in greatest need, are not served. Relation-extraction datasets are another source family: the gold set of HeRE \citep{keinan2026here} could be recast as NLI with templates, as in English \citep{sainz-etal-2021-label}. Where it is available, conversion takes days rather than the months of expert annotation an NLI corpus demands. We invite the community to build such resources, to publish the diagnostic alongside them, and to report the negative cases as we have.

\section{Annotation Guidelines}
\label{app:guidelines}

The instructions given to both annotators, condensed. Operational directions (file handling, row ordering) are omitted.

\paragraph{Label set.} Exactly one of four values per item: \emph{entailment}, the hypothesis must be true given the premise; \emph{contradiction}, it must be false, both cannot hold; \emph{neutral}, it may be true or false and the premise does not settle it; and \emph{incoherent}, the hypothesis is not well-formed Hebrew so its meaning cannot be judged.

\paragraph{Decision rules.} Judge from the premise text alone, without outside knowledge. A paraphrase of the premise counts as entailment however different the wording. A plausible but unstated detail is neutral: ``sounds likely'' is not entailment. A hypothesis that reverses or negates a fact the premise states explicitly is a contradiction, even where a single word differs. Where the logic is clear but the Hebrew is broken, mark it incoherent rather than guessing at the intended meaning.

\paragraph{Pronouns.} A pronoun in the hypothesis is read as referring to an entity in the premise, both sentences being taken to describe the same situation. This follows the convention of SNLI and MultiNLI, which arises from SNLI's photo-caption collection protocol rather than from any written rule; an annotation study of the related SICK corpus found problems in its annotation guidelines \citep{kalouli-etal-2019-explaining}. Where the referent is genuinely ambiguous, for instance several candidate entities in the premise, the item is neutral: the annotator does not guess a referent.

\paragraph{Independence.} Annotators worked alone and did not compare answers before both files were returned, and neither had access to the automatic labels at any point.

\paragraph{Potential annotation bias.} The contradiction rule accepts a single-token reversal of an explicitly stated fact. That is standard NLI practice, but single-token edits are common among the corpus's contradictions, so the instruction predisposes annotators to accept them. It should be read alongside the per-annotator spread on that class in Appendix~\ref{app:human}.

\section{Human Validation, Per Source}
\label{app:human}

All 315 items, 63 per source. \emph{IAA} is agreement between the annotators on items both labeled; A1 and A2 are each annotator's agreement with the pipeline's label. Items an annotator marked incoherent are left out of that annotator's figures.

\begin{table}[h]
\centering\small
\begin{tabular}{lccc}
\toprule
\textbf{Source} & \textbf{IAA} & \textbf{A1} & \textbf{A2} \\
\midrule
HeQ            & 0.950 & 0.983 & 0.968 \\
QASem          & 0.793 & 0.780 & 0.839 \\
HeSum          & 0.817 & 0.817 & 0.855 \\
HebSumm.\      & 0.810 & 0.746 & 0.873 \\
HebPara.\   & 0.855 & 0.823 & 0.919 \\
\midrule
\textbf{All}   & \textbf{0.845} & \textbf{0.829} & \textbf{0.891} \\
\bottomrule
\end{tabular}
\caption{Agreement by source. Cohen's $\kappa$ \citep{cohen1960coefficient}: 0.763 between annotators ($n = 303$); 0.742 (A1) and 0.836 (A2) against the automatic labels.}
\label{tab:humanfull}
\end{table}

HeQ is the easiest source for both annotators; QASem and the summarization sources are the hardest, and both annotators agree least with the automatic labels there.

\paragraph{Details.} Annotator 1 marked 11 items incoherent and annotator 2 marked 4. On the 303 items with two NLI labels the annotators agree on 84.5\% of items, and on the 256 items where they agree the automatic labels reach $\kappa = 0.910$. The second judge (Qwen3.6-27B) scores $\kappa = 0.69$ and $0.62$ against the two annotators and $0.75$ against their consensus on the 315 items. On the 2{,}000 sampled contradictions it agrees with the pipeline on 90.7\% of those grounded in a detail the premise states and 87.3\% of the rest, ranging from 77\% to 98\% across sources and the two groups.

\section{All Judges on the 315 Items}
\label{app:zeroshot}

Table~\ref{tab:zeroshotfull} scores every judge on the same 315 items by source and by gold label, and gives each judge's agreement with annotator 2. Both LLMs are prompted in English over the Hebrew pair at temperature 0; seven of gemini's answers could not be mapped to a label and are left out of its figures. Items a judge marked incoherent or answered unparsably are left out of its row in Table~\ref{tab:zeroshot} ($n$ = 304, 311, 308, 315).

\begin{table}[h]
\centering\small
\setlength{\tabcolsep}{4pt}
\resizebox{\columnwidth}{!}{%
\begin{tabular}{lcccc}
\toprule
& \textbf{A1} & \textbf{A2} & \textbf{gemini} & \textbf{gpt-5.4} \\
\midrule
\multicolumn{5}{l}{\emph{vs.\ the automatic label, by source}} \\
HeQ            & 0.983 & 0.968 & 0.952 & 0.905 \\
QASem          & 0.780 & 0.839 & 0.797 & 0.746 \\
HeSum          & 0.817 & 0.855 & 0.714 & 0.714 \\
HebSumm.\      & 0.746 & 0.873 & 0.651 & 0.635 \\
HebPara.\   & 0.823 & 0.919 & 0.885 & 0.683 \\
\midrule
\multicolumn{5}{l}{\emph{vs.\ the automatic label, by gold label}} \\
entailment     & 0.875 & 0.914 & 0.874 & 0.952 \\
neutral        & 0.962 & 0.943 & 0.952 & 0.800 \\
contradiction  & 0.632 & 0.812 & 0.560 & \textbf{0.457} \\
\midrule
accuracy       & 0.829 & 0.891 & 0.799 & 0.737 \\
$\kappa$       & 0.742 & 0.836 & 0.697 & 0.605 \\
\midrule
\multicolumn{5}{l}{\emph{vs.\ annotator 2}} \\
agreement      & 0.845 & --- & 0.770 & 0.714 \\
$\kappa$       & 0.763 & --- & 0.648 & 0.565 \\
\bottomrule
\end{tabular}}
\caption{All four judges on the 315 items (gemini: \texttt{gemini-3.5-flash-lite}; gpt-5.4: \texttt{gpt-5.4-mini}).}
\label{tab:zeroshotfull}
\end{table}

Contradiction separates the judges: agreement on that class runs from 0.812 (annotator 2) to 0.457 (gpt-5.4-mini), while entailment and neutral stay at 0.80 to 0.96 for every judge.

\section{Dataset Statistics}
\label{app:data}

Table~\ref{tab:datastats} gives, per source, the original corpus size, the converted pairs per label and the splits. Each training condition subsamples the train split, class-balanced: 3{,}000 pairs per source (1{,}000 per label), 15{,}000 for the pooled \textsc{all}, 15{,}000 for HebNLI-15k and 5{,}838 for LCHAIM-max.

\begin{table}[h]
\centering\footnotesize
\setlength{\tabcolsep}{3pt}
\resizebox{\columnwidth}{!}{%
\begin{tabular}{lrrrrrrr}
\toprule
& & \multicolumn{3}{c}{\textbf{converted pairs}} & \multicolumn{3}{c}{\textbf{splits}} \\
\cmidrule(lr){3-5}\cmidrule(lr){6-8}
\textbf{Source} & \textbf{Corpus} & \textbf{Ent.} & \textbf{Con.} & \textbf{Neu.} & \textbf{Train} & \textbf{Dev} & \textbf{Test} \\
\midrule
HeQ         & 30{,}147 & 21{,}784 & 9{,}156 & 21{,}784 & 47{,}505 & 2{,}580 & 2{,}639 \\
QASem       & 20{,}000 & 21{,}157 & 3{,}128 & 21{,}070 & 42{,}852 &   781 & 1{,}722 \\
HeSum       & 10{,}000 &  9{,}997 & 3{,}897 &  9{,}997 & 19{,}072 & 2{,}382 & 2{,}437 \\
HebSumm.    &  3{,}547 &  5{,}024 & 4{,}328 &  4{,}952 & 10{,}111 & 2{,}091 & 2{,}102 \\
HebPara. & 9{,}792 & 17{,}382 & 9{,}958 &  9{,}846 & 29{,}655 & 3{,}818 & 3{,}713 \\
\midrule
\multicolumn{2}{l}{\emph{translated benchmarks}} & & & & & & \\
HebNLI      & --- & --- & --- & --- & 300{,}067 & 1{,}999 & 884 \\
LCHAIM      & --- & --- & --- & --- &  6{,}719 &   799 & 805 \\
\bottomrule
\end{tabular}}
\caption{Per-source statistics: original corpus size, converted pairs per label, and released splits.}
\label{tab:datastats}
\end{table}

\section{Experimental Setup Details}
\label{app:setup}

\subsection{Hyperparameters}

Table~\ref{tab:hparams} gives the training configuration for both architectures. Decoders use LoRA \citep{hu-etal-2022-lora} with a paged 8-bit AdamW optimizer \citep{dettmers2023qlora, dettmers2022optimizers}. Larger decoders use a smaller per-device batch with more gradient accumulation, and Qwen3.6-27B's adapter targets the attention and MLP projections.

\begin{table}[h]
\centering\small
\setlength{\tabcolsep}{3pt}
\resizebox{\columnwidth}{!}{%
\begin{tabular}{lcc}
\toprule
& \textbf{Encoders} & \textbf{Decoders} \\
\midrule
Objective        & 3-way clf.\ head & LoRA causal-LM SFT \\
Max sequence len & 2048 & 2048 \\
Learning rate    & 2e-5 & 1e-4 \\
Optimizer        & AdamW & paged AdamW (8-bit) \\
Warmup ratio     & 0.10 & 0.03 \\
Weight decay     & 0.01 & 0.0 \\
Batch size       & 32 & 8 \\
Grad.\ accum.    & 1 & 2 \\
Max epochs       & 20 & 3 \\
Max steps (cap)  & 8000 & 8000 \\
Min steps        & 200 & 100 \\
Evals per epoch  & 4 & 4 \\
Early-stop patience & 5 evals & 5 evals \\
Model selection  & dev macro-F1 & dev loss \\
\midrule
LoRA rank $r$    & --- & 16 \\
LoRA $\alpha$    & --- & 32 \\
LoRA dropout     & --- & 0.05 \\
LoRA targets     & --- & all-linear \\
Precision        & --- & bf16 \\
Max new tokens   & --- & 8 \\
\bottomrule
\end{tabular}}
\caption{Training configuration. Decoder loss covers only the target tokens; sequence length is clamped to each model's limit (512 for XLM-R and AlephBERT).}
\label{tab:hparams}
\end{table}

\subsection{Decoder Prompt and Target}

Decoders are trained and evaluated with an instruction prompt naming the three permitted labels, with the loss restricted to the target tokens:

\begin{samepage}
\begin{quote}\small\ttfamily
[Hebrew instruction: ``Below are two sentences. Determine the logical relation\\
between sentence 1 and sentence 2 and answer in one word only:\\
entailment, contradiction, or neutral.'']\\
Sentence 1: \{premise\}\\
Sentence 2: \{hypothesis\}\\
Answer:
\end{quote}
\end{samepage}

Targets are the three Hebrew label words (\textit{hesek}, \textit{stira}, \textit{neytrali}). Evaluation is greedy generation followed by label normalization, which strips reasoning blocks (\texttt{<think>}\ldots\texttt{</think>}) before matching. Over-long \emph{premises} are truncated from the end so the hypothesis and answer cue always survive; this matters because HeSum, HebSummaries and LCHAIM premises are full articles.

Two per-model settings are set via a per-model flag table rather than globally: wrapping the prompt in the tokenizer's own chat template (required by the gemma-4 family, which otherwise continues the prompt as raw text instead of emitting a label), and zeroing \texttt{token\_type\_ids} for encoders with \texttt{type\_vocab\_size}${=}1$. Decoders are evaluated with vLLM \citep{kwon2023efficient}, except fine-tuned Qwen3.6-27B, whose adapter vLLM cannot load; it is evaluated through Hugging Face PEFT \citep{peft}.

\subsection{Hardware}

Experiments ran on a SLURM cluster (NVIDIA B200 nodes, 4-hour wall clock), an 8$\times$A100-80GB host, and V100 nodes for some encoder runs. Jobs killed at the wall clock resume from their latest checkpoint.

\section{Full Results}
\label{app:results}

\subsection{Per-Model Results}
\label{app:heatmaps}\label{app:permodel}\label{app:lchaim}

Tables~\ref{tab:indomain}--\ref{tab:permodel_lcahim} give every model's mean and seed SD: in-domain, on HebNLI and on LCHAIM, by training condition. On LCHAIM, HebNLI-15k raises decoders by 0.074 on average and LCHAIM-max lowers their HebNLI score by 0.034, so the two translated benchmarks transfer only partly to each other.

\begin{table*}[t]\centering\scriptsize
\setlength{\tabcolsep}{4pt}
\begin{tabular}{lcccccc}
\toprule
\textbf{Model} & \textbf{HeQ} & \textbf{QASem} & \textbf{HeSum} & \textbf{HebSumm.} & \textbf{HebPara.} & \textbf{\textsc{all}} \\
\midrule
DictaLM-3.0-1.7B & .969\,{\tiny$\pm$.003} & .930\,{\tiny$\pm$.008} & .922\,{\tiny$\pm$.003} & .925\,{\tiny$\pm$.004} & .966\,{\tiny$\pm$.004} & .956 \\
DictaLM-3.0-24B & .990\,{\tiny$\pm$.001} & .957\,{\tiny$\pm$.005} & .963\,{\tiny$\pm$.006} & .968\,{\tiny$\pm$.001} & .990\,{\tiny$\pm$.002} & .976 \\
Hebatron & .982\,{\tiny$\pm$.003} & .927\,{\tiny$\pm$.004} & .929\,{\tiny$\pm$.009} & .934\,{\tiny$\pm$.010} & .981\,{\tiny$\pm$.000} & .964 \\
gemma-4-12B & .985\,{\tiny$\pm$.002} & .937\,{\tiny$\pm$.004} & .906\,{\tiny$\pm$.032} & .833\,{\tiny$\pm$.091} & .983\,{\tiny$\pm$.006} & .938 \\
gemma-4-31B & .993\,{\tiny$\pm$.001} & .952\,{\tiny$\pm$.004} & .963\,{\tiny$\pm$.002} & .942\,{\tiny$\pm$.020} & .994\,{\tiny$\pm$.001} & .974 \\
Qwen3.6-27B & .993\,{\tiny$\pm$.002} & .950\,{\tiny$\pm$.008} & .956\,{\tiny$\pm$.004} & .948\,{\tiny$\pm$.008} & .993\,{\tiny$\pm$.002} & .968 \\
\midrule
AlephBERT & .855\,{\tiny$\pm$.016} & .805\,{\tiny$\pm$.011} & .722\,{\tiny$\pm$.024} & .652\,{\tiny$\pm$.030} & .743\,{\tiny$\pm$.071} & .832 \\
NeoDictaBERT & .545\,{\tiny$\pm$.342} & .691\,{\tiny$\pm$.307} & .345\,{\tiny$\pm$.014} & .622\,{\tiny$\pm$.003} & .894\,{\tiny$\pm$.005} & .761 \\
XLM-R-large & .963\,{\tiny$\pm$.006} & .888\,{\tiny$\pm$.019} & .844\,{\tiny$\pm$.018} & .606\,{\tiny$\pm$.033} & .912\,{\tiny$\pm$.027} & .908 \\
XLM-R-XL & .966\,{\tiny$\pm$.003} & .917\,{\tiny$\pm$.013} & .867\,{\tiny$\pm$.005} & .757\,{\tiny$\pm$.075} & .865\,{\tiny$\pm$.131} & .913 \\
mE5-large & .751\,{\tiny$\pm$.356} & .897\,{\tiny$\pm$.001} & .383\,{\tiny$\pm$.067} & .621\,{\tiny$\pm$.059} & .898\,{\tiny$\pm$.015} & .891 \\
\bottomrule
\end{tabular}
\caption{In-domain macro-F1 (train and test on the same source), mean $\pm$ seed SD. \textsc{all}: the pooled model, averaged over the five test sets.}
\label{tab:indomain}
\end{table*}

\begin{table*}[t]\centering\scriptsize
\setlength{\tabcolsep}{3pt}
\resizebox{\textwidth}{!}{%
\begin{tabular}{lccccccccc}
\toprule
\textbf{Model} & \textbf{ZS} & \textbf{HeQ} & \textbf{QASem} & \textbf{HeSum} & \textbf{HebSumm.} & \textbf{HebPara.} & \textbf{\textsc{all}} & \textbf{HebNLI-15k} & \textbf{LCHAIM-max} \\
\midrule
DictaLM-3.0-1.7B & .514 & .667\,{\tiny$\pm$.035} & .665\,{\tiny$\pm$.031} & .458\,{\tiny$\pm$.089} & .530\,{\tiny$\pm$.019} & .680\,{\tiny$\pm$.002} & .630\,{\tiny$\pm$.014} & .860\,{\tiny$\pm$.003} & .533\,{\tiny$\pm$.078} \\
DictaLM-3.0-24B & .632 & .780\,{\tiny$\pm$.026} & .728\,{\tiny$\pm$.028} & .542\,{\tiny$\pm$.052} & .671\,{\tiny$\pm$.042} & .697\,{\tiny$\pm$.017} & .669\,{\tiny$\pm$.050} & .900\,{\tiny$\pm$.007} & .660\,{\tiny$\pm$.122} \\
Hebatron & .649 & .737\,{\tiny$\pm$.026} & .760\,{\tiny$\pm$.011} & .503\,{\tiny$\pm$.025} & .596\,{\tiny$\pm$.047} & .746\,{\tiny$\pm$.017} & .712\,{\tiny$\pm$.055} & .902\,{\tiny$\pm$.010} & .680\,{\tiny$\pm$.025} \\
gemma-4-12B & .756 & .750\,{\tiny$\pm$.050} & .768\,{\tiny$\pm$.014} & .621\,{\tiny$\pm$.011} & .750\,{\tiny$\pm$.019} & .797\,{\tiny$\pm$.021} & .730\,{\tiny$\pm$.027} & .898\,{\tiny$\pm$.011} & .751\,{\tiny$\pm$.056} \\
gemma-4-31B & .863 & .836\,{\tiny$\pm$.014} & .820\,{\tiny$\pm$.020} & .695\,{\tiny$\pm$.014} & .749\,{\tiny$\pm$.003} & .842\,{\tiny$\pm$.028} & .768\,{\tiny$\pm$.029} & .917\,{\tiny$\pm$.003} & .758\,{\tiny$\pm$.086} \\
Qwen3.6-27B & .760 & .827\,{\tiny$\pm$.034} & .796\,{\tiny$\pm$.011} & .705\,{\tiny$\pm$.031} & .739\,{\tiny$\pm$.042} & .865\,{\tiny$\pm$.015} & .731\,{\tiny$\pm$.042} & .912\,{\tiny$\pm$.008} & .589\,{\tiny$\pm$.077} \\
\midrule
AlephBERT & --- & .255\,{\tiny$\pm$.038} & .379\,{\tiny$\pm$.011} & .230\,{\tiny$\pm$.020} & .219\,{\tiny$\pm$.044} & .314\,{\tiny$\pm$.006} & .356\,{\tiny$\pm$.006} & .672\,{\tiny$\pm$.016} & .327\,{\tiny$\pm$.044} \\
NeoDictaBERT & --- & .343\,{\tiny$\pm$.057} & .467\,{\tiny$\pm$.115} & .280\,{\tiny$\pm$.018} & .274\,{\tiny$\pm$.008} & .415\,{\tiny$\pm$.015} & .376\,{\tiny$\pm$.034} & .812\,{\tiny$\pm$.010} & .306\,{\tiny$\pm$.013} \\
XLM-R-large & --- & .319\,{\tiny$\pm$.030} & .400\,{\tiny$\pm$.014} & .261\,{\tiny$\pm$.071} & .231\,{\tiny$\pm$.091} & .330\,{\tiny$\pm$.033} & .419\,{\tiny$\pm$.004} & .407\,{\tiny$\pm$.328} & .170\,{\tiny$\pm$.005} \\
XLM-R-XL & --- & .513\,{\tiny$\pm$.018} & .532\,{\tiny$\pm$.044} & .244\,{\tiny$\pm$.008} & .410\,{\tiny$\pm$.068} & .357\,{\tiny$\pm$.027} & .462\,{\tiny$\pm$.042} & .827\,{\tiny$\pm$.007} & .235\,{\tiny$\pm$.116} \\
mE5-large & --- & .401\,{\tiny$\pm$.094} & .509\,{\tiny$\pm$.048} & .292\,{\tiny$\pm$.049} & .333\,{\tiny$\pm$.051} & .384\,{\tiny$\pm$.054} & .437\,{\tiny$\pm$.022} & .780\,{\tiny$\pm$.012} & .254\,{\tiny$\pm$.057} \\
\bottomrule
\end{tabular}}
\caption{Macro-F1 on HebNLI by training condition, mean $\pm$ seed SD. ZS: zero-shot (decoders only).}
\label{tab:permodel_hebnli}
\end{table*}

\begin{table*}[t]\centering\scriptsize
\setlength{\tabcolsep}{3pt}
\resizebox{\textwidth}{!}{%
\begin{tabular}{lccccccccc}
\toprule
\textbf{Model} & \textbf{ZS} & \textbf{HeQ} & \textbf{QASem} & \textbf{HeSum} & \textbf{HebSumm.} & \textbf{HebPara.} & \textbf{\textsc{all}} & \textbf{HebNLI-15k} & \textbf{LCHAIM-max} \\
\midrule
DictaLM-3.0-1.7B & .272 & .407\,{\tiny$\pm$.023} & .378\,{\tiny$\pm$.020} & .320\,{\tiny$\pm$.034} & .359\,{\tiny$\pm$.024} & .346\,{\tiny$\pm$.004} & .396\,{\tiny$\pm$.011} & .448\,{\tiny$\pm$.029} & .586\,{\tiny$\pm$.036} \\
DictaLM-3.0-24B & .443 & .523\,{\tiny$\pm$.020} & .397\,{\tiny$\pm$.020} & .441\,{\tiny$\pm$.027} & .491\,{\tiny$\pm$.021} & .321\,{\tiny$\pm$.088} & .449\,{\tiny$\pm$.035} & .551\,{\tiny$\pm$.064} & .701\,{\tiny$\pm$.006} \\
Hebatron & .471 & .540\,{\tiny$\pm$.011} & .487\,{\tiny$\pm$.009} & .439\,{\tiny$\pm$.034} & .496\,{\tiny$\pm$.037} & .533\,{\tiny$\pm$.037} & .460\,{\tiny$\pm$.049} & .592\,{\tiny$\pm$.021} & .655\,{\tiny$\pm$.034} \\
gemma-4-12B & .608 & .534\,{\tiny$\pm$.027} & .529\,{\tiny$\pm$.018} & .482\,{\tiny$\pm$.024} & .561\,{\tiny$\pm$.048} & .550\,{\tiny$\pm$.065} & .478\,{\tiny$\pm$.030} & .634\,{\tiny$\pm$.007} & .696\,{\tiny$\pm$.008} \\
gemma-4-31B & .701 & .625\,{\tiny$\pm$.001} & .548\,{\tiny$\pm$.029} & .557\,{\tiny$\pm$.033} & .586\,{\tiny$\pm$.016} & .676\,{\tiny$\pm$.011} & .535\,{\tiny$\pm$.026} & .673\,{\tiny$\pm$.020} & .768\,{\tiny$\pm$.013} \\
Qwen3.6-27B & .621 & .560\,{\tiny$\pm$.046} & .564\,{\tiny$\pm$.005} & .574\,{\tiny$\pm$.039} & .516\,{\tiny$\pm$.029} & .634\,{\tiny$\pm$.052} & .517\,{\tiny$\pm$.047} & .663\,{\tiny$\pm$.007} & .749\,{\tiny$\pm$.018} \\
\midrule
AlephBERT & --- & .327\,{\tiny$\pm$.012} & .298\,{\tiny$\pm$.024} & .298\,{\tiny$\pm$.012} & .285\,{\tiny$\pm$.034} & .310\,{\tiny$\pm$.007} & .328\,{\tiny$\pm$.015} & .323\,{\tiny$\pm$.001} & .330\,{\tiny$\pm$.036} \\
NeoDictaBERT & --- & .299\,{\tiny$\pm$.023} & .281\,{\tiny$\pm$.015} & .322\,{\tiny$\pm$.022} & .265\,{\tiny$\pm$.012} & .202\,{\tiny$\pm$.019} & .327\,{\tiny$\pm$.014} & .188\,{\tiny$\pm$.019} & .324\,{\tiny$\pm$.019} \\
XLM-R-large & --- & .322\,{\tiny$\pm$.005} & .338\,{\tiny$\pm$.021} & .287\,{\tiny$\pm$.013} & .281\,{\tiny$\pm$.019} & .259\,{\tiny$\pm$.073} & .324\,{\tiny$\pm$.006} & .230\,{\tiny$\pm$.080} & .259\,{\tiny$\pm$.060} \\
XLM-R-XL & --- & .341\,{\tiny$\pm$.006} & .289\,{\tiny$\pm$.014} & .277\,{\tiny$\pm$.008} & .310\,{\tiny$\pm$.013} & .213\,{\tiny$\pm$.029} & .326\,{\tiny$\pm$.011} & .247\,{\tiny$\pm$.014} & .271\,{\tiny$\pm$.170} \\
mE5-large & --- & .292\,{\tiny$\pm$.047} & .310\,{\tiny$\pm$.006} & .236\,{\tiny$\pm$.061} & .316\,{\tiny$\pm$.007} & .228\,{\tiny$\pm$.017} & .330\,{\tiny$\pm$.009} & .201\,{\tiny$\pm$.004} & .280\,{\tiny$\pm$.012} \\
\bottomrule
\end{tabular}}
\caption{Macro-F1 on LCHAIM by training condition, mean $\pm$ seed SD.}
\label{tab:permodel_lcahim}
\end{table*}

\subsection{Translated to Native, Per Model}
\label{app:revbreak}

Table~\ref{tab:revfull} gives each model's scores on the five \textsc{HEntailment} test sets after HebNLI-15k fine-tuning. The three decoders with the highest zero-shot scores lose on all five sources and send most neutral items to contradiction; the weak decoders gain on most sources.

\begin{table*}[t]\centering\scriptsize
\setlength{\tabcolsep}{4pt}
\begin{tabular}{lcccccc}
\toprule
\textbf{Model} & \textbf{HeQ} & \textbf{QASem} & \textbf{HeSum} & \textbf{HebSumm.} & \textbf{HebPara.} & \textbf{neu$\to$con} \\
\midrule
DictaLM-3.0-1.7B & .658 ({\tiny +0.200}) & .718 ({\tiny +0.283}) & .437 ({\tiny +0.082}) & .356 ({\tiny $-$0.013}) & .634 ({\tiny +0.383}) & 0.06$\to$0.30 \\
DictaLM-3.0-24B & .629 ({\tiny +0.017}) & .676 ({\tiny +0.140}) & .510 ({\tiny +0.180}) & .303 ({\tiny $-$0.018}) & .582 ({\tiny +0.055}) & 0.14$\to$0.30 \\
Hebatron & .666 ({\tiny $-$0.074}) & .717 ({\tiny +0.134}) & .495 ({\tiny +0.058}) & .458 ({\tiny +0.027}) & .496 ({\tiny $-$0.092}) & 0.19$\to$0.40 \\
gemma-4-12B & .578 ({\tiny $-$0.283}) & .651 ({\tiny $-$0.076}) & .364 ({\tiny $-$0.309}) & .220 ({\tiny $-$0.389}) & .466 ({\tiny $-$0.263}) & 0.02$\to$0.75 \\
gemma-4-31B & .739 ({\tiny $-$0.222}) & .763 ({\tiny $-$0.049}) & .426 ({\tiny $-$0.417}) & .185 ({\tiny $-$0.462}) & .643 ({\tiny $-$0.300}) & 0.03$\to$0.59 \\
Qwen3.6-27B & .646 ({\tiny $-$0.288}) & .687 ({\tiny $-$0.052}) & .417 ({\tiny $-$0.355}) & .204 ({\tiny $-$0.371}) & .436 ({\tiny $-$0.358}) & 0.03$\to$0.67 \\
\midrule
AlephBERT & .424 & .566 & .288 & .307 & .509 & --- \\
NeoDictaBERT & .360 & .706 & .232 & .246 & .432 & --- \\
XLM-R-large & .364 & .393 & .232 & .297 & .368 & --- \\
XLM-R-XL & .544 & .706 & .259 & .357 & .583 & --- \\
mE5-large & .235 & .697 & .243 & .243 & .429 & --- \\
\bottomrule
\end{tabular}
\caption{Models fine-tuned on HebNLI-15k, on the five \textsc{HEntailment} test sets (macro-F1, mean of three seeds; in brackets, change against zero-shot). \emph{neu$\to$con}: share of gold-neutral items labeled contradiction, zero-shot $\to$ fine-tuned.}
\label{tab:revfull}
\end{table*}

\subsection{Low-Resource Initialization}
\label{app:lowres}

Table~\ref{tab:lowres} gives the full low-resource curve of \S\ref{sec:results}.

\begin{table}[t]\centering\scriptsize
\setlength{\tabcolsep}{3pt}
\begin{tabular}{lrccrc}
\toprule
\textbf{Model} & $N$ & \textbf{scratch} & \textbf{init.} & \textbf{gain} & \textbf{seeds$>$0} \\
\midrule
NeoDictaBERT & 100 & .336\,{\tiny$\pm$.020} & .543\,{\tiny$\pm$.022} & +0.207 & 3/3 \\
 & 500 & .380\,{\tiny$\pm$.016} & .625\,{\tiny$\pm$.039} & +0.245 & 3/3 \\
 & 1k & .381\,{\tiny$\pm$.018} & .694\,{\tiny$\pm$.040} & +0.313 & 3/3 \\
 & 5k & .731\,{\tiny$\pm$.020} & .758\,{\tiny$\pm$.007} & +0.027 & 3/3 \\
 & 15k & .812\,{\tiny$\pm$.010} & .811\,{\tiny$\pm$.007} & $-$0.001 & 1/3 \\
\midrule
XLM-R-large & 100 & .398\,{\tiny$\pm$.023} & .643\,{\tiny$\pm$.008} & +0.245 & 3/3 \\
 & 500 & .398\,{\tiny$\pm$.237} & .733\,{\tiny$\pm$.006} & +0.335 & 3/3 \\
 & 1k & .601\,{\tiny$\pm$.281} & .746\,{\tiny$\pm$.002} & +0.146 & 1/3 \\
 & 5k & .773\,{\tiny$\pm$.013} & .766\,{\tiny$\pm$.009} & $-$0.008 & 1/3 \\
 & 15k & .407\,{\tiny$\pm$.328} & .802\,{\tiny$\pm$.005} & +0.395 & 3/3 \\
\midrule
DictaLM-3.0-24B & 100 & .849\,{\tiny$\pm$.020} & .801\,{\tiny$\pm$.071} & $-$0.048 & 1/3 \\
 & 500 & .866\,{\tiny$\pm$.019} & .878\,{\tiny$\pm$.013} & +0.011 & 2/3 \\
 & 1k & .877\,{\tiny$\pm$.005} & .876\,{\tiny$\pm$.008} & $-$0.001 & 1/3 \\
 & 5k & .897\,{\tiny$\pm$.015} & .893\,{\tiny$\pm$.008} & $-$0.005 & 1/3 \\
 & 15k & .900\,{\tiny$\pm$.007} & .900\,{\tiny$\pm$.002} & $-$0.000 & 2/3 \\
\midrule
gemma-4-31B & 100 & .908\,{\tiny$\pm$.005} & .892\,{\tiny$\pm$.002} & $-$0.015 & 0/3 \\
 & 500 & .909\,{\tiny$\pm$.004} & .907\,{\tiny$\pm$.008} & $-$0.002 & 1/3 \\
 & 1k & .911\,{\tiny$\pm$.002} & .912\,{\tiny$\pm$.009} & +0.002 & 2/3 \\
 & 5k & .909\,{\tiny$\pm$.003} & .912\,{\tiny$\pm$.007} & +0.002 & 2/3 \\
 & 15k & .917\,{\tiny$\pm$.003} & .911\,{\tiny$\pm$.007} & $-$0.006 & 1/3 \\
\bottomrule
\end{tabular}
\caption{Low-resource HebNLI: $N$ examples, from scratch or initialized from pooled \textsc{HEntailment}; HebNLI macro-F1, mean $\pm$ seed SD.}
\label{tab:lowres}
\end{table}

\subsection{Robustness Checks}
\label{app:robust}

\paragraph{Balanced re-scoring.} Each run is re-scored on label-balanced subsamples of every test set. The mean change is $+0.000$ (HeQ), $+0.026$ (QASem), $+0.004$ (HeSum), $-0.002$ (HebSummaries), $-0.005$ (Hebrew-Paraphrase), $-0.002$ (HebNLI) and $-0.003$ (LCHAIM), and cell rankings are preserved ($\rho \geq 0.987$).

\paragraph{AFLite.} Adversarial filtering \citep{lebras2020adversarial} with mE5-large probes removes 40\% to 67\% of each source's test rows. On the remaining items, grid scores change by $+0.000$ (HeQ), $+0.025$ (QASem), $+0.015$ (HeSum), $-0.011$ (HebSummaries) and $-0.003$ (Hebrew-Paraphrase) on average; the largest single-run change is 0.12.

\paragraph{Verified test subset.} Section~\ref{sec:validation} reports the overall confirmation rate (87.8\%). By label, Qwen3.6-27B confirms 91\% of entailments, 86\% of contradictions and 85\% of neutrals. It disagrees most with HebSummaries neutrals (51\% confirmed, the rest called contradictions) and QASem entailments (70\%, mostly called neutral). Fine-tuning on HebNLI-15k changes every decoder's score in the same direction on the confirmed items and on the full test set. Qwen3.6-27B, which chose the items, gains the most from the subset (+0.050 zero-shot).

\end{document}